\documentclass[letterpaper]{article} 
\usepackage{aaai2026}  
\usepackage{times}  
\usepackage{helvet}  
\usepackage{courier}  
\usepackage[hyphens]{url}  
\usepackage{graphicx} 
\usepackage{natbib}  
\usepackage{caption} 
\usepackage{algorithm}

\usepackage{newfloat}
\usepackage{listings}

\usepackage{amssymb}
\usepackage{amsmath} 
\usepackage{booktabs}
\usepackage{enumerate}
\usepackage{graphicx}
\usepackage{subfigure}
\usepackage{xspace}
\usepackage{float}
\usepackage{bm}
\usepackage{multirow}
\usepackage{booktabs}
\usepackage{color}
\usepackage{framed}
\usepackage{iitem}
\usepackage[T1]{fontenc}
\definecolor{shadecolor}{RGB}{180,180,180}
\usepackage{url}
\newcommand{\ie}{\emph{i.e.,}\xspace}

\newcommand{\eg}{\emph{e.g.,}\xspace}

\newcommand{\ignore}[1]{}

\usepackage{algpseudocode}
\usepackage{amsthm}
\usepackage{amsmath}
\usepackage{tikz}

\usepackage{amsmath}
\usepackage{colortbl}
\usepackage{color, xcolor}

\usepackage{amsthm}
\usepackage{amssymb}
\usepackage{amsmath}  
\usepackage{braket}   
\usepackage{booktabs}
\usepackage{bm}
\usepackage[table]{xcolor}
\usepackage{multirow}
\usepackage{graphicx}
\usepackage{subcaption}
\usepackage{algpseudocode}

\DeclareCaptionStyle{ruled}{labelfont=normalfont,labelsep=colon,strut=off} 
\floatstyle{ruled}
\newfloat{listing}{tb}{lst}{}
\floatname{listing}{Listing}
\title{L2V-CoT: Cross-Modal Transfer of Chain-of-Thought Reasoning \\ via Latent Intervention}
\author{
    Yuliang Zhan\equalcontrib,
    Xinyu Tang\equalcontrib,
    Han Wan,
    Jian Li,
    Ji-Rong Wen,
    Hao Sun\thanks{Corresponding author.}
}

\affiliations{
    Gaoling School of Artificial Intelligence, Renmin University of China, Beijing, China\\

    \{zhanyuliang, xinyun\_tang, wanhan2001, lijian2022, jrwen, haosun\}@ruc.edu.cn
}

\usepackage{bibentry}

\begin{document}

\maketitle

\begin{abstract} 
Recently, Chain-of-Thought (CoT) reasoning has significantly enhanced the capabilities of large language models (LLMs), but Vision–Language Models (VLMs) still struggle with multi-step reasoning tasks due to limited multimodal reasoning data. To bridge this gap, researchers have explored methods to transfer CoT reasoning from LLMs to VLMs. However, existing approaches either need high training costs or require architectural alignment. In this paper, we use Linear Artificial Tomography (LAT) to empirically show that LLMs and VLMs share similar low-frequency latent representations of CoT reasoning despite architectural differences. Based on this insight, we propose \textbf{L2V-CoT}, a novel training-free latent intervention approach that transfers CoT reasoning from LLMs to VLMs. \textbf{L2V-CoT} extracts and resamples low-frequency CoT representations from LLMs in the frequency domain, enabling dimension matching and latent injection into VLMs during inference to enhance reasoning capabilities.  Extensive experiments demonstrate that our approach consistently outperforms training-free baselines and even surpasses supervised methods. Our code is available at \url{https://github.com/whynot-zyl/L2V-CoT/}.
\end{abstract}


\section{Introduction}

Recently, reasoning models have achieved significant advancements in tackling a wide range of complex tasks~\cite{openai2024learningtoreason,wang2025bee,guo2025deepseek,wang2025unveiling}.
A crucial factor contributing to this success is the Chain-of-Thought~(CoT) method, which allows large language models~(LLMs) to decompose complex problems into a sequence of intermediate reasoning steps~\cite{xia2024beyond}.
This approach has significantly improved their reasoning and decision-making capabilities~\cite{zhang2025enhancing}.
In contrast, although Vision–Language Models~(VLMs) have shown impressive results on tasks such as visual question answering~\cite{hartsock2024vision} and image captioning~\cite{kim2024structure,li2025analyzing}, they still struggle with tasks that require multi-step reasoning, such as chart and geometric analysis~\cite{cheng2025comt,zhang2024mathverse}. 
This problem is primarily attributed to the scarcity of multimodal reasoning data, as generating such data is both resource-intensive and time-consuming~\cite{liu2025we}.

To address this problem, recent studies have explored ways to transfer reasoning capabilities from LLMs to VLMs. 
Virgo~\cite{du2025virgo} enables cross-modal reasoning transfer by training VLMs on large amounts of textual Chain-of-Thought~(CoT) data.
However, this approach is hindered by high training costs and limited generalization ability.
To overcome this limitation, existing studies transfer reasoning capabilities from LLM to VLM through model merging~\cite{chen2025bring,zhan2024over}.
Although this method achieves effective transfer, its applicability is limited to scenarios where the VLM and the source LLM are architecturally aligned.
However, in real-world scenarios, the LLM backbone of VLM is not always aligned with the strong text reasoning model, which limits the upper bound of reasoning capabilities that can be transferred from LLMs to VLMs.
This raises a practical challenge: How can we transfer reasoning abilities from LLMs to VLMs across \textbf{different architectures}?

Inspired by Contrast-Consistent Search~\cite{burns2022discovering}, which suggests that model capabilities can be captured and manipulated through linear transformations, we explore the possibility of transferring reasoning capabilities across different model architectures.
However, since models with different modalities often have distinct internal structures, it remains unclear whether the internal reasoning patterns encoded in one modality can be effectively interpreted, aligned, and transferred to another modality.

To further analyze reasoning capabilities across models of different modalities, we apply Linear Artificial Tomography~(LAT)~\cite{zou2023representation}, a representation reading method that extracts latent states with contrastive inputs, to examine the latent representations of them.
The findings of our analysis reveal that:
(1) The \textbf{low-frequency components} of VLM’s CoT direction representations, derived from linear modeling of CoT and Non-CoT representations, can activate its reasoning ability. In contrast, the high-frequency components do not help.
(2) These low-frequency representations have a \textbf{similar distribution} to those of LLMs. 
The observation suggests a consistent structural alignment in the latent space across modalities, which enables effective cross-modal transfer of reasoning ability.
Based on these insights, we propose \underline{\textbf{L}}atent Intervention for \underline{\textbf{L}}LM-to-\underline{\textbf{V}}LM \underline{\textbf{CoT}} Transfertion (\textbf{L2V-CoT}), a training-free latent intervention method that transfers the general CoT reasoning capabilities of LLMs to VLMs.
Specifically, we collect CoT and Non-CoT from LLMs to construct contrastive samples. 
These samples are then encoded by the reasoning LLM to obtain CoT direction representations.
However, representations derived from different architectures often encounter the dimension mismatch problem.
To address this, we apply low-pass filtering to the CoT direction representations from the LLM to preserve essential CoT information.
Then, we perform resampling in the frequency domain to match the dimension.
These resampled representations are injected into the VLM via latent intervention during inference, thereby implicitly enhancing its reasoning ability.
As a training-free and model-agnostic method, L2V-CoT enables efficient transfer the reasoning capabilities of LLMs to VLMs.
To evaluate its effectiveness, we conduct experiments on multiple visual reasoning benchmarks across diverse VLMs. 
The experimental results demonstrate that L2V-CoT consistently outperforms other training-free baselines and even surpasses some supervised approaches.

Our contributions can be summarized as follows:

$\bullet$ To our best knowledge, we are the first to leverage Linear Artificial Tomography to analyze the transferability of reasoning capabilities between LLMs and VLMs.

$\bullet$ We propose L2V-CoT, a novel training-free method that transfers the general CoT reasoning capability of LLMs to VLMs, thereby enhancing the reasoning ability of VLMs.

$\bullet$ Extensive experiments validate the effectiveness of our approach in transferring CoT reasoning across modalities.
\section{Related Work}

\paragraph{Multimodal Chain-of-Thought Reasoning.}
As CoT reasoning proves effective in LLMs, recent work has extended it to multimodal tasks~\cite{chen-etal-2024-m3cot, li2025unleashing}. Existing methods fall into two main categories: explicit and implicit methods, which are orthogonal and complementary in enhancing VLM reasoning~\cite{wu2025boosting}.
Explicit methods guide reasoning step-by-step via rewards or search without altering model states~\cite{yao2024mulberry}. 
In contrast, implicit methods improve reasoning by modifying internal states~\cite{luo2025ursa,zhang2023multimodal}. LlamaV-o1~\cite{thawakar2025llamav} learns from multimodal CoT data. Given the scarcity of visual CoT annotations, Virgo~\cite{du2025virgo} trains VLMs on textual CoT data~\cite{du2025virgo}. Parameter mergingmethods further enable training-free transfer from LLMs to VLMs but are constrained to architecture-aligned models~\cite{chen2025bring}.
Our method, L2V-CoT, belongs to the implicit category.  It enables architecture-agnostic, training-free CoT transfer from LLMs to VLMs via latent intervention, offering a flexible and generalizable solution for multimodal reasoning.

\paragraph{Activation engineering.}
Activation engineering modifies a model’s latent states for two main purposes: (1) understanding internal mechanisms and (2) controlling behavior~\cite{zou2023representation}. It uses representation reading to identify latent states tied to high-level concepts, supporting both interpretability and intervention. This method been applied to reducing hallucinations~\cite{li2023inference,tang2025enhancing,li2024images} and adjusting sentiment~\cite{hollinsworth-etal-2024-language}.
While recent work has explored using activation engineering for capability transfer within the model~\cite{tang2025unlocking}, its potential in cross-modal reasoning remains underexplored. In this work, we extend it to understanding and transferring CoT reasoning capabilities across modalities.
\section{Preliminary}
\paragraph{Linear Artificial Tomography.}

Linear Artificial Tomography (LAT) is a representation reading technique for identifying internal representations of high-level concepts in deep neural networks~\cite{zou2023representation}. 
It serves as a tool for analyzing model behavior, which operates through a three-step process: (1) Designing Stimulus and Task, (2) Collecting Neural Activity, and (3) constructing a Linear Model. 

First, to activate the model’s internal representation associated with a target concept or function $f$, LAT convert prompt $\{p_i\}_{i=1}^n$ into positive $\{p_i^+\}_{i=1}^n$ (elicits $f$) and negative $\{p_i^-\}_{i=1}^n$ (does not). Next, $p_i^+$ and $p_i^-$ are fed into the model $\mathcal{M}$ to extract hidden states $h_i^+(l)$ and $h_i^-(l)$ from layer $l$. 
In this paper, We extracts representations from final-token hidden states~\cite{tang2025unlocking}. Finally, LAT constructs a linear model (Direction Represent) $u$.
The mean of $\{u_i\}_{i=1}^{n}$ can be used as a transparent and interpretable method for analyzing or activating the concept.

\section{Empirical Analysis}
\label{sec:empiricalanalisis}
In this section, We first introduce the use of LAT in this work. We empirically analyze how VLM and LLM encode CoT reasoning apability using LAT.
\paragraph{Inducing and Capturing Internal Activations.}

To analyze the transferability of reasoning capabilities across modalities, we use LAT to observe the internal representations of LLMs and VLMs. 
Specifically, we first design two types of prompts: $\{q_i^+\}_{i=1}^n$ (\ie "Let’s think step by step.") and $\{q_i^-\}_{i=1}^n$ (\ie "Answer the question directly").
These prompts are separately fed into the LLM to generate the CoT responses $\{c\}_{i=1}^n$ and the non-CoT responses $\{d\}_{i=1}^n$. Given positive inputs $\{c_i\}_{i=1}^n$ and negative inputs $\{d_i\}_{i=1}^n$, we extract CoT representation sets $\{h_L(c_i,l)\}_{i=1}^n$ and $\{h_V(c_i,l)\}_{i=1}^n$, as well as vanilla representation sets $\{h_L(d_i,l)\}_{i=1}^n$ and $\{h_V(d_i,l)\}_{i=1}^n$,
where $h_L$ and $h_V$ denote internal representations of the LLM and VLM.
For our implementation, we use LLaMA3-8B~\cite{meta2024introducing} to encode LLM representations and Qwen2-VL-7B-Instruct~\cite{wang2024qwen2} to encode VLM representations. 
\begin{figure}[t!]
    \centering
    \includegraphics[width=\linewidth]{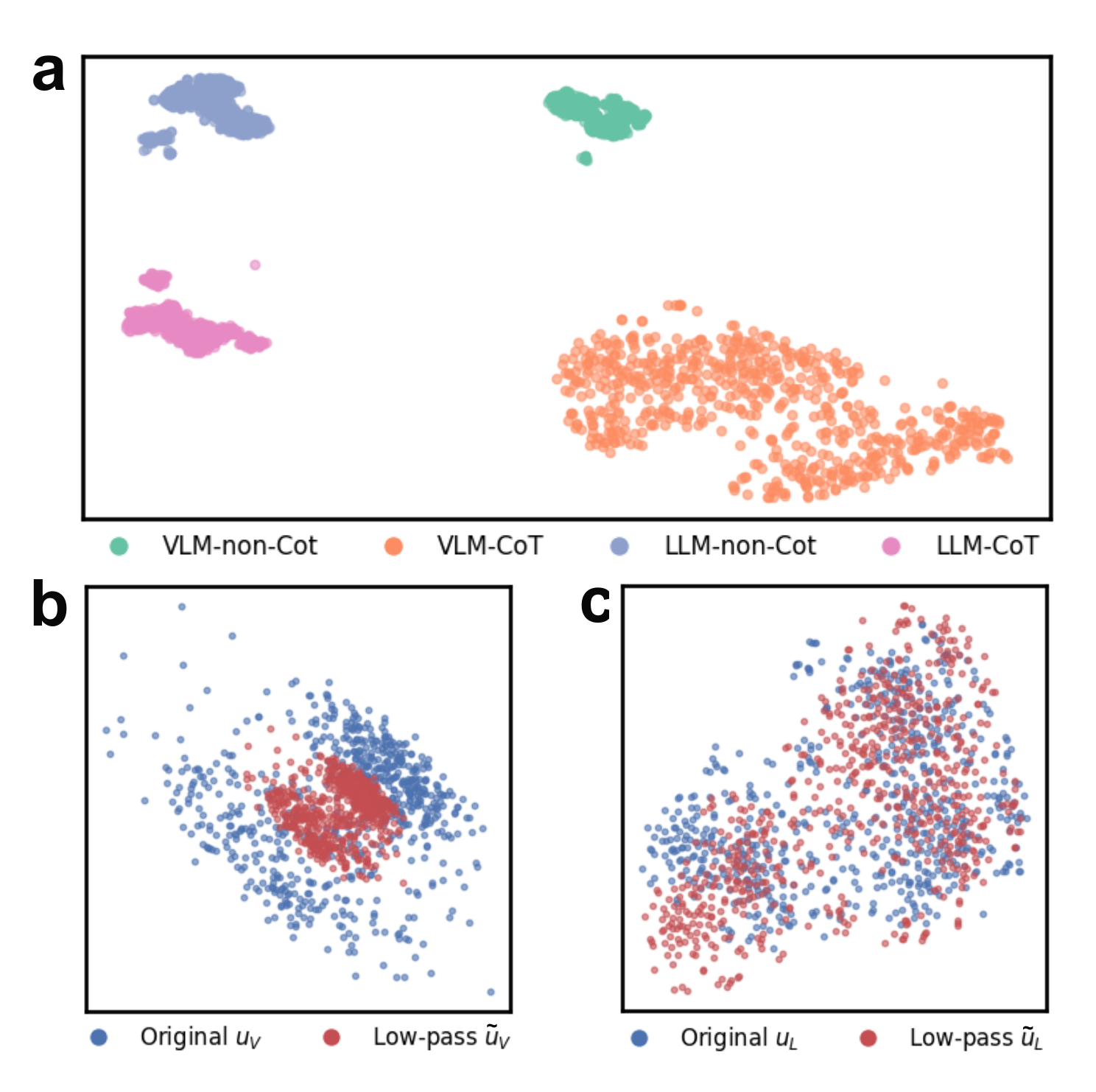}
    \vspace{-15pt}
    \caption{
        (a) Distribution of CoT and non-CoT representations in VLMs and LLMs. 
        (b) Effect of low-pass filtering on VLM CoT direction representation. 
        (c) Effect of low-pass filtering on LLM CoT direction representation.
    }
    \vspace{-8pt}
    \label{fig:distribution}
\end{figure}

\paragraph{VLMs and LLMs have similar CoT reasoning encoding patterns.}
To better understand the properties of CoT representations and non-CoT representations within both LLMs and VLMs, we apply PCA~\cite{jolliffe2002springer} for dimensionality reduction and visualize the 2D representations.
The visualization results are presented in Figure~\ref{fig:distribution}a.
We observe that both VLMs and LLMs produce tightly clustered CoT representations across samples, which occupy distinct regions in the latent space compared to their non-CoT counterparts.
These findings suggest that VLM and LLM have similar representational encoding schemes for CoT reasoning, despite their architectural differences. 
This consistency raises an question: \textit{Can we leverage this shared encoding pattern to effectively transfer CoT reasoning capabilities from LLMs to VLMs?}
However, we observe that the CoT representations of VLM cluster less tightly than LLM.
The observed discrepancy is due to the heterogeneity between visual and textual modalities: when visual inputs are introduced and jointly trained with language, they induce representation drift in the high-dimensional latent space~\cite{yangadapting}. 
This drift emerges from parameter updates during multimodal joint training and persists in the representations even when processing unimodal textual inputs~\cite{parkgeneralizing}.
Consequently, this results in a gap between the CoT direction representations of LLMs and VLMs, which we will examine in detail in the following analysis.
\paragraph{Low-Frequency CoT direction representations of VLM encode CoT features and activate reasoning ability.}
We establish linear models to extract the CoT direction representation set $u_V(l)$ and $u_L(l)$ for VLM and LLM, respectively. Formally, these are defined as:
\begin{align}
    \label{eq:delta_cot}
    u_L(l) &= \{h_L(c_i,l) - h_L(d_i,l)\}_{i=1}^n, \\
    u_V(l) &= \{h_V(c_i,l) - h_V(d_i,l)\}_{i=1}^n,
\end{align}
where $h(c_i, l)$ and $h(d_i, l)$ are hidden states at layer $l$ corresponding to the CoT and non-CoT responses, computed by VLM ($h_V$) or LLM ($h_L$). 
To measure the dispersion of these representations, we compute the trace of their covariance matrix, where a larger trace indicates higher dispersion.
\begin{equation}
    \label{eq:covariance_trace}
    \mathrm{Tr}(u(l)) = \frac{1}{n} \sum_{i=1}^n \left\| u_i - \bar{u} \right\|^2,
\end{equation}
where \( \bar{u} = \frac{1}{n} \sum_{i=1}^n u_i \), and \( \{u_i\}_{i=1}^n \) is drawn from either \( u_L(l) \) or \( u_V(l) \).
We found that VLMs show significantly higher dispersion (1117.8) than LLMs (176.7). 
This divergence arises from the fundamental differences in architecture and training methods~\cite{schroditwo}.
LLM capabilities are encoded through linearly separable structures in activation space~\cite{burns2022discovering},
In contrast, VLMs have distinct optimization processes during multimodal pre-training.
Consequently, components such as cross-modal attention, projection fusion, and alignment losses are incompatible with the LLM backbone.
These inconsistencies introduce a more nonlinear and noisy activation space, ultimately causing shifts in internal representations of VLMs~\cite{yangadapting}.
\begin{figure}[t!]
    \centering
    \includegraphics[width=\linewidth]{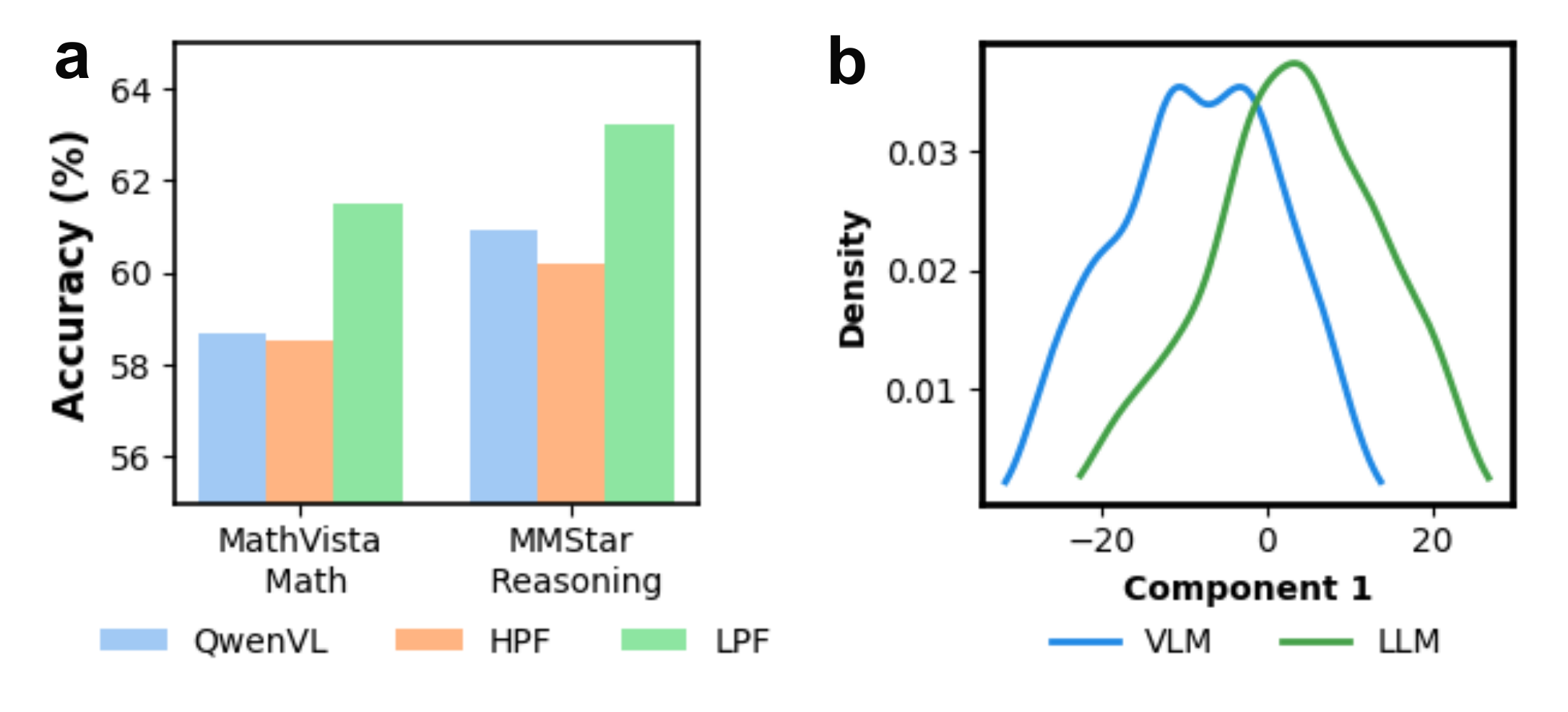}
    \vspace{-15pt}
    \caption{{(a)} Performance of Qwen2VL and its injected variants on MathVista-math and MMStar-reasoning. “HPF” injects high-frequency features. “LPF” injects low-frequency features. {(b)} The direction representation distribution for VLM and LLM after low-pass filtering(math domain).}
    \label{fig:same}
    \vspace{-15pt}
\end{figure}
\begin{figure*}[t!]
    \centering
    \includegraphics[width=\linewidth]{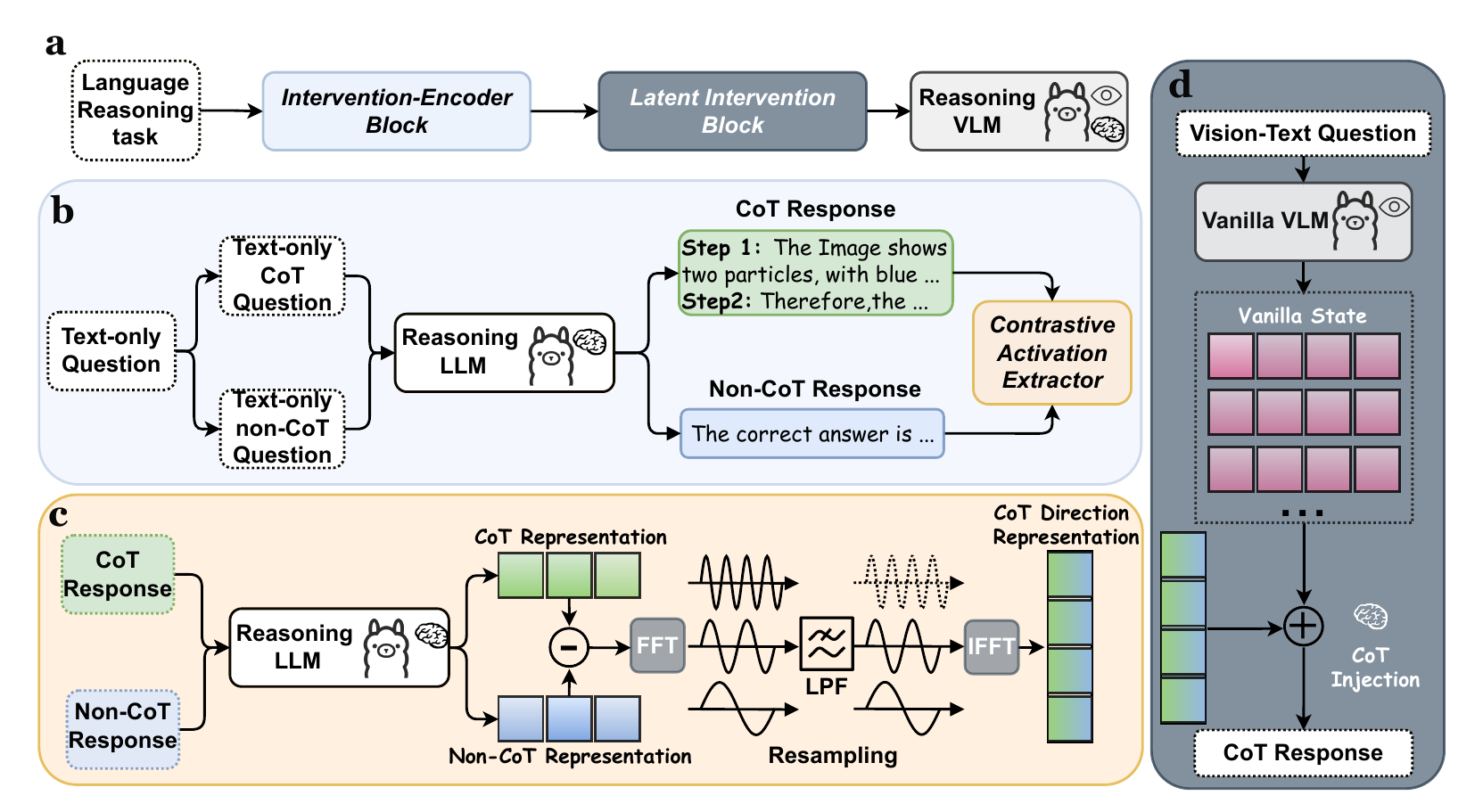}
    \vspace{-15pt}
    \caption{{(a)} The overview of Latent Intervention for LLM-to-VLM CoT Transfertion (L2V-CoT). {(b)} The block of Intervention-Encoder. {(c)} Low-pass LLM CoT direction representation extraction process. {(d)} The block of Latent Intervention.
    }
    \label{fig:main}
    \vspace{-8pt}
\end{figure*}
To gain a deeper understanding of  CoT direction representation shifts, we apply Fourier-domain low-pass filtering to the representations of VLMs, obtaining the low-frequency CoT direction representation $\tilde{u}_V(l)$:
\begin{equation}
    \label{eq:VLM_lowpass}
        \tilde{u}_V(l) = \text{Re} \left[\text{IFFT} \left( \mathbf{M_k} \odot \text{FFT}(u_V(l)) \right)\right],
\end{equation}
where $\text{FFT}(\cdot)$ denotes the Fast Fourier Transform, $\text{IFFT}(\cdot)$ is its inverse operation, $\odot$ represents Hadamard product and $\text{Re}[\cdot]$ extracts the real part to eliminate residual imaginary components. The low-pass mask $\mathbf{M_k} \in {0,1}^d$, where $d$ is the domain of $u_V(l)$, retains the first $k$ frequency components and is defined as follows:
\begin{equation}
    \label{eq:low-pass}
        \mathbf{M}_k[i] = 
        \begin{cases}
        1, & \text{if } i < \frac{k}{2} \text{ or } i > d - \frac{k}{2}, \\
        0, & \text{otherwise}.
        \end{cases}
\end{equation}
Figure~\ref{fig:distribution}b shows the distribution of the original $u_V(l)$ and filtered representations $\text{low-pass } \tilde{u}_V$. 
We can observe that filtered representations become significantly more concentrated, and the distribution shape stays stable.
Quantitatively, trace drops from 1117.8 to 197.7 (close to LLM's 176.7).
For comparison, applying the same low-pass filtering to ${u}_L(l)$ just causes minimal change (Figure~\ref{fig:distribution}c). And we get the trace of $\text{low-pass } \tilde{u}_L$ is 169.3, only slightly lower than the original value of 176.7. 
This indicates that filtering removes visual-specific noise while preserving LLM reasoning capability.

To verify if low-frequency filtering preserves CoT information, we average low and high-frequency VLM direction representations and inject them into the VLM. We conduct experiments on the MathVista and MMStar datasets (detailed in the Experiments section). The results are shown in Figure~\ref{fig:same}. We observe that only low-frequency direction representation enhances the reasoning ability of VLM. This further shows that low-frequency CoT direction representations in VLMs carry CoT information and can be used to enhance the reasoning ability of model.

\paragraph{Distribution of Low-Frequency CoT direction representations in VLM and LLM exhibit substantial overlap. }
Due to the scarcity of multimodal reasoning data, the reasoning ability of VLMs falls far behind that of LLMs.
Although low-frequency CoT directions can enhance the VLM reasoning abilities, they still limited by the capacity of the VLM itself.
\textit{To overcome this bottleneck, we investigate transferring the superior reasoning capabilities of LLMs to VLMs via latent intervention.}
To support this, we analyze the alignment between low-frequency CoT direction representations of VLMs and LLMs.
As shown in Figure~\ref{fig:same}b, their PCA-projected 1D distributions exhibit substantial overlap, indicating a shared latent structure. 
This indicates the potential for effective knowledge transfer without significant domain mismatch. 
Consequently, this empirical observation provides strong evidence supporting the feasibility of transferring reasoning capabilities across modalities and disparate model architectures through latent interventions.

\section{CoT Reasoning Capability transfer via \\Latent Intervention}
\label{sec:method}
As previously discussed, LLM and VLM have similar low-pass CoT direction representations. This suggests that the CoT capabilities of LLMs can be transferred to VLMs. We extract CoT direction representations from LLM and inject them to VLMs. However, due to representation dimension mismatch between different architectures, we apply low-pass filtering and resampling in the latent space to align dimension while minimizing the loss of CoT ability. Figure~\ref{fig:main} shows the framework. In this section, we first describe the extraction of low-pass CoT pattern representations and then introduce how these direction representations are used to perform latent intervention on vanilla VLMs.

\begin{table*}[t!]
\centering
\scalebox{0.77}{
\begin{tabular}{c|c|ccc|cccccc|ccc|c|c|c}
\hline
\multirow{2}{*}{\textbf{Model}} & \multirow{2}{*}{\textbf{Method}} & \multicolumn{3}{c|}{\textbf{MathVista}} & \multicolumn{6}{c|}{\textbf{MathVerse Benchmarks}} & \multicolumn{3}{c|}{\textbf{MMStar}} & \multirow{2}{*}{\textbf{DM}} & \multirow{2}{*}{\textbf{MV}} & \multirow{2}{*}{\textbf{Avg.}} \\
\cline{3-14}
& & All & General & Math & Overall & T-D & T-L & V-I & V-D & V-O & All & Percep. & Reason. & & & \\
\hline
\multirow{7}{*}{\textbf{LLAVA}} 
& Non-CoT response         & 35.2 & 46.7 & 25.3 & 20.9 & 24.0 & 20.9 & 21.5 & 20.0 & 18.2 & 43.4 & 52.8 & 38.7  & 22.9 & 12.8 & 27.0 \\
& Few-Shot CoT            & 33.7 & 45.8  & 23.4  & 19.3 & 22.8 & 19.3 & 19.7 & 18.4 & 16.5 & 40.4 & 49.5 & 35.9  & 21.1 & 11.5 & 25.2 \\
& MathNeuro               & 35.8 & 46.6 & 26.6 & 21.5 & 24.6 & 21.5 & 21.7 & 20.8 & 18.9 & 44.1 & 51.3 & 40.5 & 23.0 & 13.4 & 27.6 \\
& Modal Merging        & 36.4 & 45.3 & 28.8  & 21.9 & 26.8 & 22.4 & 22.9 & 20.8 & 16.6 & 43.6 & 51.3 & 39.8  & 23.5 & 13.4 & 27.8 \\
& RoT                     & 36.8 & 47.3  & 27.9  & 22.9 & 25.8 & 23.3 & 23.1 & 22.0 & 20.5 & 45.6 & 53.1 & 41.9  & 24.8 & 14.8 & 29.0 \\
& Finetuned CoT           & 39.9 & 49.1  & 32.0  & 24.1 & 27.6 &   24.8 & 24.0 & 22.6 & 21.6 & 47.0 & 54.4 & 43.3  & 25.8 & 15.6 & 30.5 \\
& L2V-COT (ours)         & \textbf{41.8} & \textbf{51.0} & \textbf{34.0} & \textbf{25.5} & \textbf{29.5} & \textbf{25.9} & \textbf{25.0} & \textbf{23.2} & \textbf{23.7} & \textbf{48.1} & \textbf{55.6} & \textbf{44.3}  & \textbf{26.9} & \textbf{16.2} & \textbf{31.6} \\
\hline
\multirow{7}{*}{\textbf{InternVL}} 
& Non-CoT response         & 59.3 & 62.8  & 56.3  & 29.9 & 30.2 & 33.4 & 30.8 & 32.1 & 23.0 & 59.5 & 58.0 & 60.3  & 30.5 & 19.6 & 39.8 \\
& Few-Shot CoT            & 57.4 & 61.2  & 54.1  & 27.9 & 28.6 & 31.1 & 28.4 & 30.2 & 21.1 & 57.8 & 56.3 & 58.5  & 28.9 & 18.2 & 38.0 \\
& MathNeuro               & 59.9 & 63.1  & 57.1  & 30.2 & 31.0 & 34.1 & 31.3 & 31.0 & 23.8 & 59.9 & 57.5 & 61.1  & 31.6 & 21.0 & 40.5 \\
& Modal Merging        & 60.4 & 63.4  & 57.8  & 30.9 & 30.3 & 34.6 & 32.5 & 33.2 & 24.1 & 59.3 & 58.5 & 59.7  & 31.9 & 21.1 & 40.7 \\
& RoT                     & 60.7 & 63.3  & 58.5  & 31.1 & 32.2 & 34.5 & 32.0 & 32.5 & 24.3 & 60.9 & 58.4 & 62.1  & 32.3 & 21.3 & 41.3 \\
& Finetuned CoT           & 61.2 & 63.5  & 59.3  & 32.6 & 35.4 & 35.3 & 33.1 & 33.6 & 25.5 & 61.6 & 59.0 & 62.9  & 33.2 & 21.9 & 42.1 \\
& L2V-COT (ours)         & \textbf{61.6} & \textbf{63.5}  & \textbf{60.0} & \textbf{33.3} & \textbf{37.2} & \textbf{35.4} & \textbf{33.9} & \textbf{34.0} & \textbf{26.1} & \textbf{62.0} & \textbf{59.1} & \textbf{63.5}  & \textbf{33.7} & \textbf{22.3} & \textbf{42.6} \\
\hline
\multirow{7}{*}{\textbf{QwenVL}} 
& Non-CoT response         & 60.5 & 62.6  & 58.7  & 26.9 & 28.6 & 27.4 & 27.8 & 29.4 & 21.1 & 63.1 & 67.4 & 60.9  & 33.8 & 19.1 & 40.7 \\
& Few-Shot CoT            & 58.4 & 61.3  & 56.0  & 25.3 & 26.5 & 26.2 & 26.4 & 27.7 & 19.5 & 60.2 & 64.1 & 58.2  & 31.9 & 17.6 & 38.7 \\
& MathNeuro               & 61.8 & 63.3  & 60.5  & 28.6 & 30.1 & 29.8 & 29.6 & 30.7 & 23.0 & 62.6 & 64.0 & 61.9  & 32.8 & 21.0 & 41.4 \\
& Modal Merging        & 61.5 & 62.3  & 60.8  & 29.9 & 32.6 & 30.3 & 29.1 & 31.9 & 25.7 & 63.0 & 66.1 & 61.5  & 34.1 & 20.3 & 41.8 \\
& RoT                     & 62.9 & 65.0  & 61.2  & 29.7 & 31.4 & 30.2 & 30.0 & 31.1 & 25.8 & 64.4 & 67.1 & 63.0  & 35.1 & 21.5 & 42.7 \\
& Finetuned CoT           & 63.7 & 65.4  & 62.3  & 32.8 & 35.1 & 33.8 & 32.4 & 34.4 & 28.2 & 65.0 & 67.9 & 63.5  & 35.3 & 22.1 & 43.8 \\
& L2V-COT (ours)         & \textbf{64.2} & \textbf{65.9}  & \textbf{62.8}  & \textbf{35.5} & \textbf{37.9} & \textbf{34.9} & \textbf{34.5} & \textbf{36.7} & \textbf{33.3} & \textbf{65.5} & \textbf{68.2} & \textbf{64.1}  & \textbf{35.9} & \textbf{22.6} & \textbf{44.7} \\
\hline
\multirow{7}{*}{\textbf{Idefics}} 
& Non-CoT response         & 48.4 & 51.7  & 45.6  & 19.8 & 22.87 & 21.85 & 20.97 & 21.2 & 12.1 & 49.2 & 54.0 & 46.8  & 25.9 & 16.5 & 32.0 \\
& Few-Shot CoT            & 46.3 & 50.2  & 43.0  & 18.1 & 21.2 & 20.0 & 19.4 & 19.6 & 10.3 & 46.8 & 51.3 & 44.5  & 24.3 & 14.9 & 30.1 \\
& MathNeuro               & 49.1 & 52.4  & 46.3  & 19.8 & 23.4 & 22.1 & 20.1 & 22.0 & 11.5 & 49.4 & 54.1 & 47.1  & 26.2 & 15.9 & 32.1 \\
& Modal Merging        & 49.3 & 52.3  & 46.8  & 21.0 & 24.5 & 23.1 & 21.5 & 22.6 & 13.1 & 49.7 & 53.2 & 47.9  & 26.3 & 16.1 & 32.5 \\
& RoT                     & 49.6 & 52.8  & 46.9  & 20.7 & 24.1 & 22.6 & 21.3 & 22.5 & 12.8 & 49.9 & 54.5 & 47.6  & 26.4 & 17.5 & 32.8 \\
& Finetuned CoT           & 50.5 & 53.4  & 48.1  & 22.3 & 25.4 & 24.2 & 22.9 & 24.6 & 14.4 & 51.4 & 55.4 & 49.4  & 28.1 & 18.6 & 34.2 \\
& L2V-COT (ours)         & \textbf{51.8} & \textbf{54.3}  & \textbf{49.7}  & \textbf{23.5} & \textbf{26.3} & \textbf{25.5} & \textbf{24.6} & \textbf{25.9} & \textbf{15.3} & \textbf{52.3} & \textbf{56.3} & \textbf{50.3}  & \textbf{29.3} & \textbf{19.2} & \textbf{35.2} \\
\hline
\end{tabular}
}
\vspace{-2pt}
\caption{Performance comparison of different methods across the MathVista (All, General, and Math-related categories), MathVerse, MMStar (All, Perception, and Reasoning categories), DynaMath (DM), and MathVision (MV) benchmarks. The best result for each task is highlighted in bold. All results are the mean values of five runs using different random seeds.}
\vspace{-8pt}
\label{table:main_result}
\end{table*}

\subsection{Extraction of low-pass CoT pattern representations}
As shown in Figure~\ref{fig:main}b and formalized in Eq.~\ref{eq:delta_cot}, we first collect positive inputs $\{c_i\}_{i=1}^n$ and negative inputs $\{d_i\}_{i=1}^n$ from an LLM with predefined questions $\{q_i\}_{i=1}^n$. These pairs are yield CoT direction representations $\{u_i(l_L)\}_{i=1}^n$ in the LLM. For unbiased transfer the general CoT reasoning capability of LLM, we compute the mean of $\{u_i(l_L)\}_{i=1}^n$ to get the CoT pattern representation ${v(l_L)}$:
\begin{equation}
    \label{eq:delta_pattern}
        v(l_L) = \frac{1}{n}\sum_{i=1}^n (h_L(c_i,l_L)-h_L(d_i,l_L)),
\end{equation}
where $l_L$ is the index of the layer from which the representation is extracted in LLM. 
Due to potentially different LLM backbones in the LLM and VLM, we must match their representation dimension.
As shown earlier, LPF can preserve CoT information in the direction representations.
Therefore, we apply LMN method~\cite{gerlach2024waveform} to resample $\{u_i(l_L)\}_{i=1}^n$ in the frequency domain.
This helps reduce the loss of CoT information during dimension alignment.
This processing is illustrated in Figure~\ref{fig:main}c. FFT and IFFT are both linear transformations. For computational efficiency, we directly apply the low-pass filter to the aggregated CoT pattern representation $v(l_L)$, obtaining the low-pass CoT pattern representation $v_{LPF}(l_L)$:
\begin{equation}
    \label{eq:delta_lowpass}
        v_{\text{LPF}}(l_L) = \text{Re} \left[ \text{IFFT} \left( \text{LMN} \left( \mathbf{M_k} \odot \text{FFT}(v(l_L)) \right)\right) \right],
\end{equation}
where LMN is resampling method in frequency domain.
Since low-pass filtering reduces the energy, we normalize the resulting vector to preserve its directional information and enhance the effectiveness of injection:
\begin{equation}
    \label{eq:low-pass-normal}
        \hat{v}_{\text{LPF}}(l_L) = \frac{||v(l_L)||_2}{||v_{\text{LPF}}(l_L)||_2}v_{\text{LPF}}(l_L).
\end{equation}

\subsection{Latent Intervention}
As shown in Figure~\ref{fig:distribution}d, we inject the low-pass CoT pattern $v_{\text{LPF}}(l_L)$ into the hidden representation $h_V(l_V)$ of the Vanilla VLM at layer $l_V$ during inference, thereby transferring the reasoning patterns from the LLM to the VLM. Here, $h_V(l_V)$ denotes the internal neural representation of the VLM activated by the vision-text input during the inference process.
To preserve the original capacity of VLM, we follow~\citet{tang2025unlocking} to normalize the updated neural activations after injecting the CoT capability, thereby reducing interference with the original representation space:

\begin{align}
    \label{eq:inject-normal}
    \hat{h}_V(l_V) &= h_V(l_V) + \alpha \cdot \hat{v}_{\text{LPF}}(l_L), \\
    \hat{h}_V(l_V) &= \frac{||{h}_V(l_V)||_2}{||\hat{h}_V(l_V)||_2}\hat{h}_V(l_V),
\end{align}
where $\alpha \in \mathbb{R}^{+}$ denotes a positive coefficient that modulates the strength of the injection.

\section{Experiments}
In this section, we first describe the experimental setup, then analyze the main experimental results, and finally conduct further analysis.
\subsection{Experimental Setup}
\paragraph{Building CoT Reasoning Samples.}
To transfer the CoT reasoning capability of LLMs, we use the open-source dataset STILL-2~\cite{min2024imitate} to extract both CoT responses and non-CoT responses.
STILL distills high-quality CoT answers and direct answers from existing large-scale reasoning models~\cite{guo2025deepseek}.
Following ~\citet{tang2025unlocking}, we randomly sample 100 examples from each of four domains: math, physics, chemistry, and biology to obtain generalizable CoT pattern representations.

\paragraph{VLMs and LLMs.}
We apply L2V-CoT to a variety of VLM architectures to evaluate the generalizability. Specifically, we conduct experiments on LLaVA-Next-LLaMA3-8B~\cite{liu2024llava}, Idefics2-8B~\cite{laurenccon2024matters}, Qwen2-VL-7B-Instruct~\cite{wang2024qwen2}, and InternVL2-8B~\cite{chen2024internvl} (abbreviated as LLaVA, Idefics, QwenVL, and InternVL).
For representation transfer, we use DeepSeek-R1-Distill-Qwen-32B~\cite{guo2025deepseek, wang2024qwen2}, a model with strong reasoning ability. During injection, we align and inject the middle layers of LLM and VLM. Details are provided in Appendix~\ref{sec:appendix_experiment_detail}.

\paragraph{Datasets.}
To assess the effectiveness of L2V-CoT, we evaluate it on five benchmark: MathVista~\cite{lu2023mathvista}, MathVerse~\cite{zhang2024mathverse}, MathVision~\cite{wang2024measuring}, Dynamath~\cite{zou2024dynamath}, and MMStar~\cite{chen2024we}. MathVerse, MathVision, and DynaMath cover algebra, geometry, and calculus, suitable task, making them well suited for evaluating the reasoning capability of model. MathVista and MMStar contain more diverse vision tasks, enabling a more comprehensive assessment of models' multifaceted abilities. Detailed can be seen in Appendix~\ref{sec:appendix_dataset}.

\paragraph{Baselines.}
To demonstrate the effectiveness of our approach, we compare it with several representative baselines: non-CoT response, CoT demonstration method (Few-shot CoT), neuron activation method (MathNeuro)~\cite{christ2024math}, representation engineering approach (RoT)~\cite{hu2024understanding}, supervised fine-tuning (Finetuned CoT)~\cite{du2025virgo}, and Model Merging~\cite{chen2025bring}. 
\subsection{L2V-CoT Substantially Improves the Reasoning Capabilities of VLMs Through Controlled Slow Thinking Injection}
\begin{figure}[t!]
    \centering
    \includegraphics[width=\linewidth]{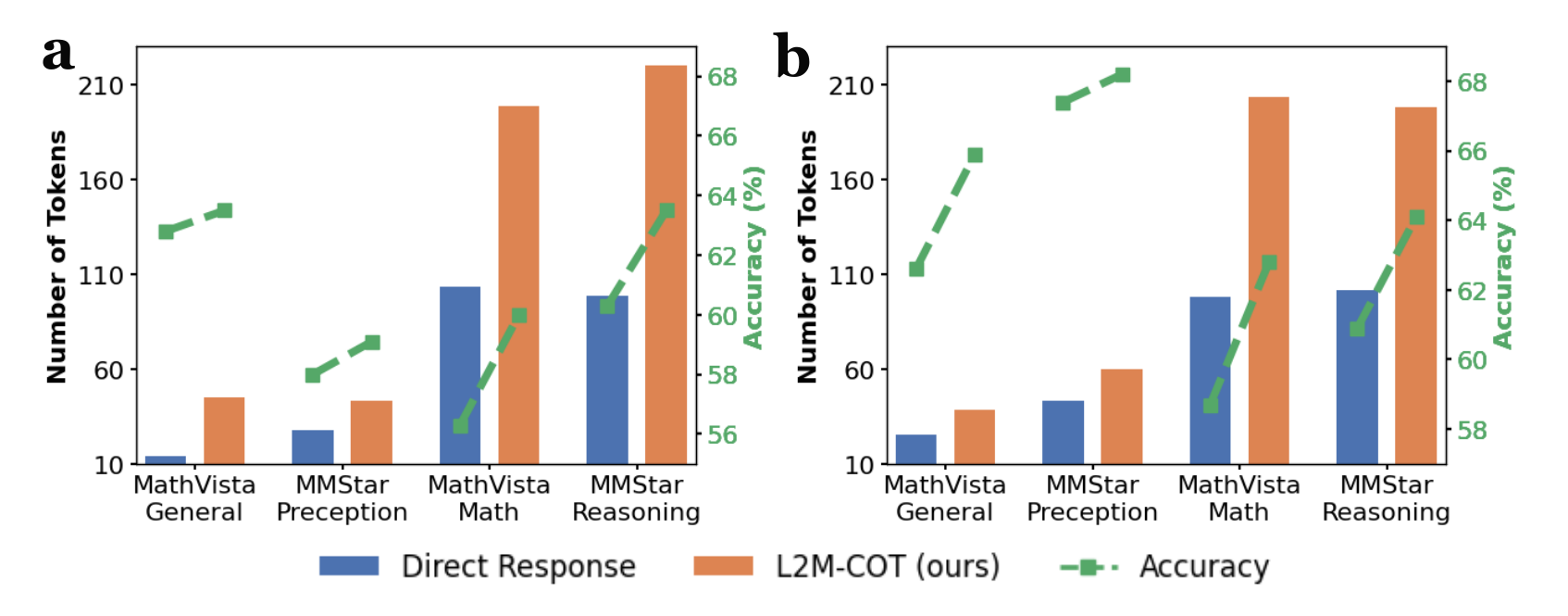}
    \vspace{-15pt}
    %
     \caption{Effect on Response Length and Accuracy. {(a)} InternVL2-8B. {(b)} Qwen2-VL-7B-Instruct.}
    \label{fig:token}
    \vspace{-8pt}
\end{figure}
The experimental results are shown in Table~\ref{table:main_result}. Few-shot CoT performs worse than non-CoT response. This is mainly because the reasoning process varies across different questions. The reasoning paths in the examples are often not generalizable and may even mislead the model’s reasoning. Unlike such explicit prompting methods, MathNeuro improves VLMs by activating reasoning-related neurons but overlooks their coordination. Modal Merging addresses this by combining a VLM and an LLM that share the same backbone. This method transfers the reasoning ability from the LLM and brings moderate improvements. However, it performs parameter-level fusion and lacks alignment between the transferred LLM and the visual modality. In contrast, RoT achieves more significant improvements. It introduces controllable reasoning activation by contrasting CoT and non-CoT prompts within the VLM's latent space. Nevertheless, RoT remains constrained by the inherent reasoning limitations of the VLM and does not facilitate the transfer of the more powerful reasoning capabilities from LLMs to VLMs.

Compared to these train-free baselines, our method achieves significantly better performance.Notably, it also outperforms models that are supervised-finetuned on datasets used to build CoT pattern representations. This improvement comes mainly from CoT pattern representations from LLM. We apply latent intervention to switch the latent space of VLM into slow-thinking mode. In this mode, the VLM is able to decompose a problem into a series of intermediate reasoning steps, which enhances its reasoning ability. A key indicator of this mode is the increased answer length. To verify that L2V-CoT transfers the slow-thinking pattern from LLMs to VLMs, we visualize the changes in answer length and accuracy on MathVista and MMStar in Figure~\ref{fig:token}. The results show that our method transfers the reasoning capability of LLMs into VLMs through latent intervention. We also observe that for tasks requiring reasoning, such as MathVista-Math and MMStar-Reasoning, the model produces significantly longer outputs and achieves greater performance improvements.

\subsection{Detailed Analysis}
\subsubsection{Layer-wise Representation Injection Universally Boosts Reasoning in Vision-Language Models.}
\begin{figure}[t!]
    \centering
    \includegraphics[width=\linewidth]{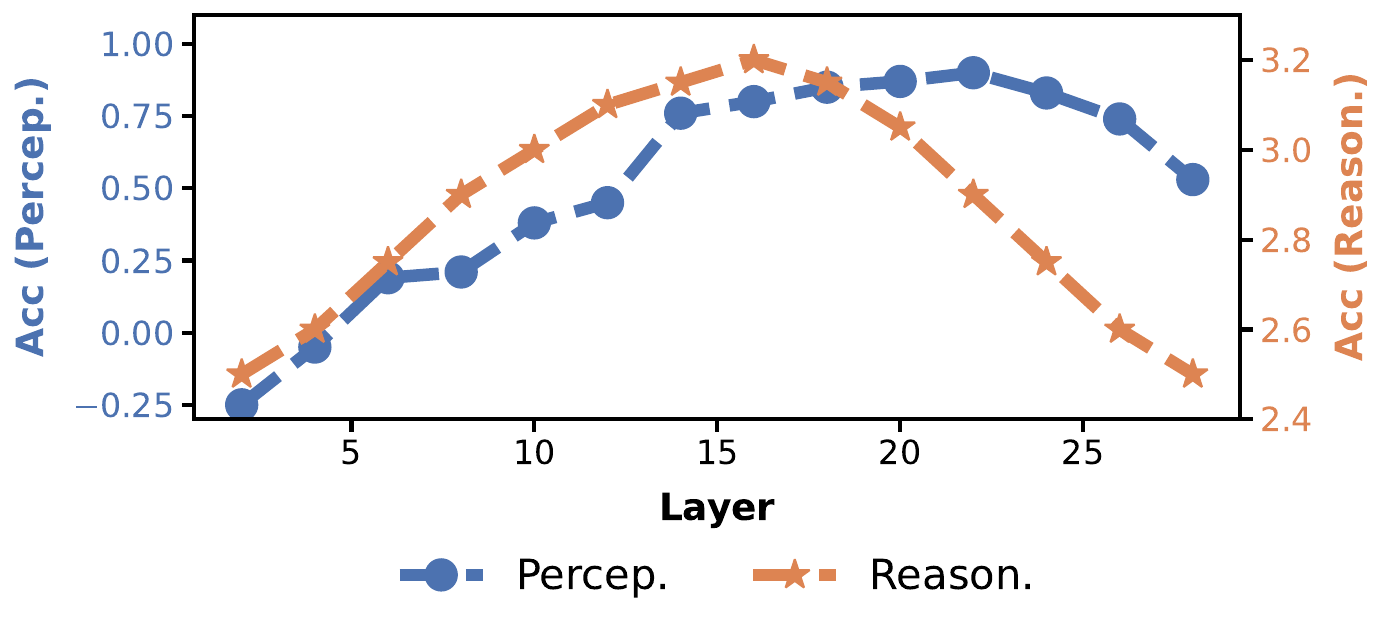}
    \vspace{-15pt}
    \caption{Qwen2-VL-7B-Instruct accuracy gain on MMStar with L2V-CoT injection at different layers. Y-axis indicates accuracy change over non-CoT responses.}
    \vspace{-8pt}
    \label{fig:token}
\end{figure}
We analyze layer-wise injection effects using Qwen2-VL-7B-Instruct, as show in Figure~\ref{fig:token}. The results show that as the injection layer increases, the performance gain first rises and then declines. This is because  lower layers encode perception features, while middle and upper layers handle reasoning~\cite{chen2025bring}. Injecting into lower layers disturbs perception. As a result, for perception tasks, injecting representations at shallow layers can even hurt performance, and the optimal injection layer tends to be deeper. In contrast, reasoning tasks benefit from early-layer injection. Additionally, when injecting into higher layers, the remaining layers are not sufficient to process the injected representations, which also leads to performance degradation. Notably, injection at any layer consistently enhances VLM reasoning.
\subsubsection{L2V-CoT Enhances Explicit Models through Complementary Reasoning Integration. }
\begin{figure}[t!]
    \centering
    \includegraphics[width=\linewidth]{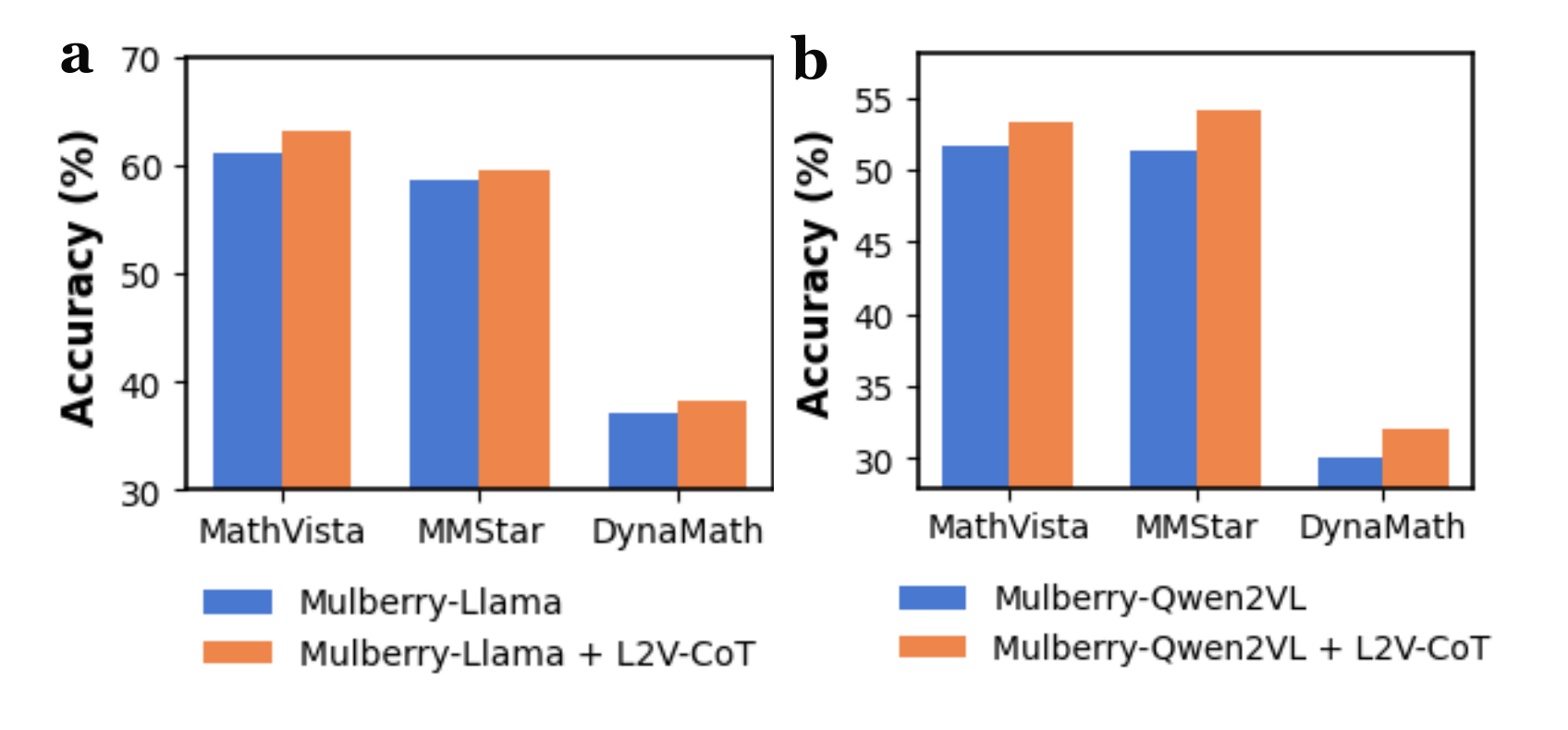}
    \vspace{-12pt}
    \caption{Performance of the explicit method \textit{Mulberry} combined with L2V-CoT. {(a)} Llama-11B. {(b)} Qwen2VL-2B.}    
    \label{fig:explicit}
\end{figure}
Explicit methods guide the reasoning process by a predefined search structure (\eg Monte Carlo Tree Search) and a reward model. In contrast, L2V-CoT changes VLM reasoning mode via latent intervention. These are orthogonal and combinable. We combine our method with the explicit method Mulberry~\cite{yao2024mulberry}, and present the results in Figure~\ref{fig:explicit}. The results show that our method leads to further performance improvements.

\subsubsection{Injecting Representations with Moderate Strength Significantly Boosts Reasoning Performance.}
\begin{figure}[t!]
    \centering
    \includegraphics[width=\linewidth]{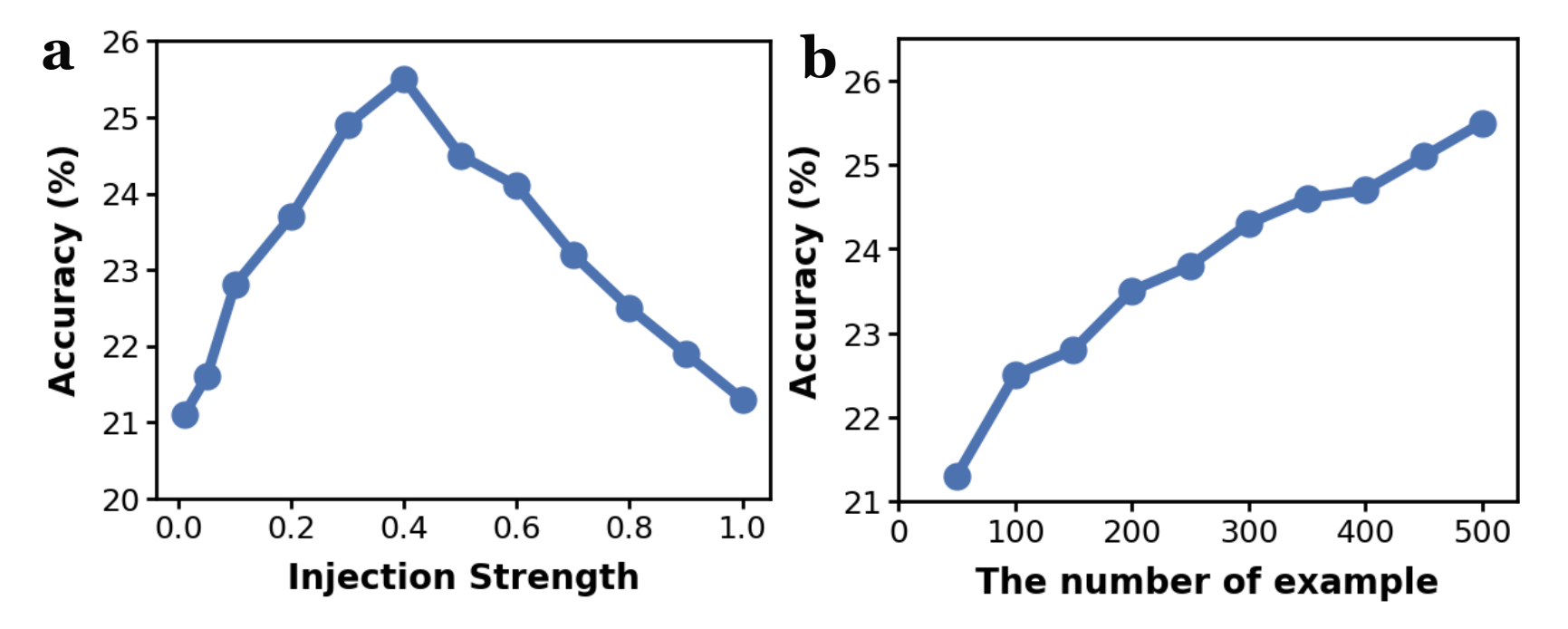}
    \vspace{-10pt}
    \caption{Effect of injection strength and CoT example number on LLaVA’s MathVerse performance (LLaVA-Next-LLaMA3-8B performance on MathVerse). {(a)} Injection strength. {(b)} Number of CoT examples.}
    \vspace{-8pt}
    \label{fig:sample_num}
\end{figure}
Injection Strength is key in our method.
We study it impact on the performance of L2V-CoT.
The results are shown in Figure~\ref{fig:sample_num}a.
When the injection strength is low, the model cannot effectively absorb the reasoning ability from the LLM.
As a result, the performance rises with increasing strength.
However, when the injection strength becomes too high, the injected representation may disrupt the original semantic information.
This can lead to a drop in performance.
Overall, representation injection consistently improves model performance.
The best performance gain is achieved when the injection strength is set to a moderate level.
\subsubsection{Performance Improves with More CoT Examples. }
L2V-CoT uses LLM CoT examples extract representation.
We investigate how example count affects performance.
The results are shown in Figure~\ref{fig:sample_num}b.
We can observe that performance improves with more examples.
This is because few examples yield CoT patterns with excessive domain bias
With more examples, L2V-CoT can extract more general and robust CoT pattern representations.
This leads to better transfer of the LLM's reasoning ability.
\subsubsection{L2V-CoT Generalizes Well Across Models of Varying Sizes.}
\begin{figure}[t!]
    \centering
    \includegraphics[width=\linewidth]{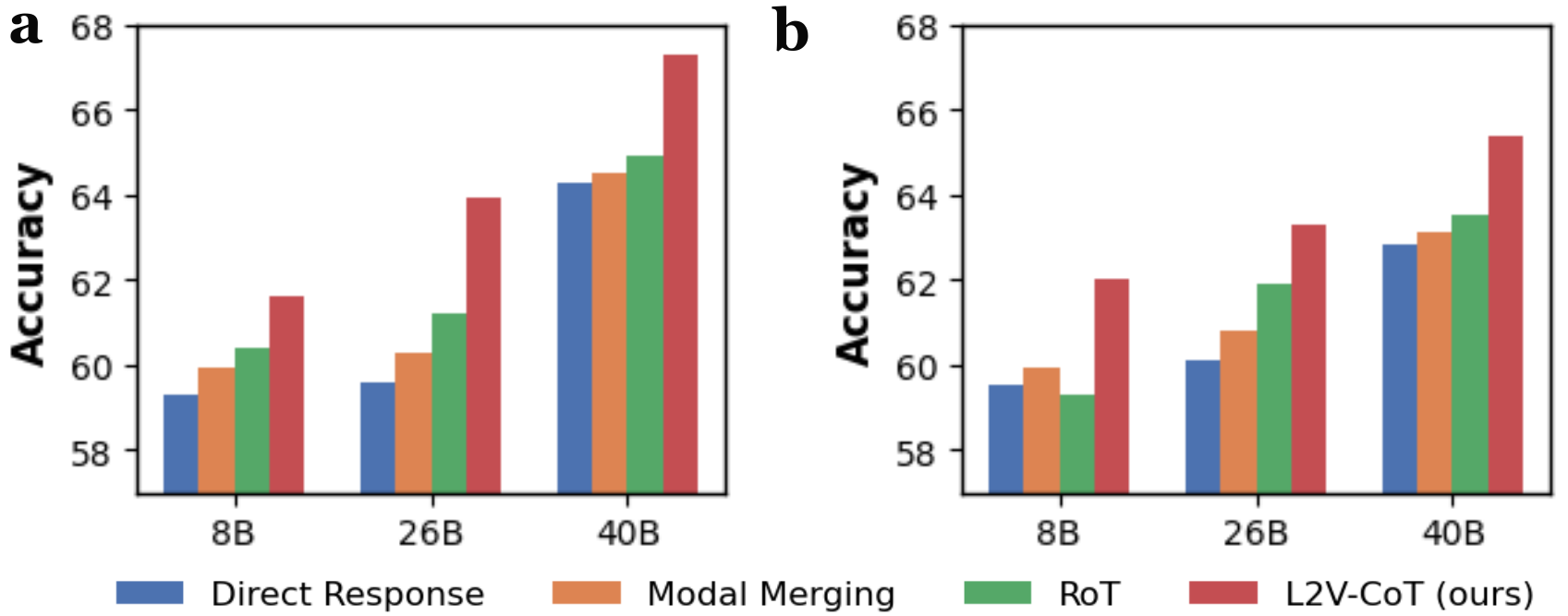}
    %
    \caption{Performance of InternVL2-series VLMs on MathVista and MMStar benchmarks. {(a)} MathVista. {(b)} MMStar.}
    \label{fig:largemodel}
\end{figure}
We test our method on InternVL2-series VLMs using MathVista and MMStar
The results are shown in Figure~\ref{fig:largemodel}.
We observe that L2V-CoT consistently achieves strong performance across models of different sizes.
This further demonstrates the effectiveness of our approach.
\subsubsection{LPF Method and LLM Representations Effectively Transfer LLM Reasoning Ability.}
\begin{table}[t!]
\centering
\scalebox{0.77}{
\begin{tabular}{c|c|c|c|c}
\hline
\textbf{Model} & \textbf{Method} & \textbf{MathVista} & \textbf{MathVerse} & \textbf{DynaMath} \\
\hline
\multirow{4}{*}{\textbf{LLAVA}} 

& Direct Response         & 35.2 & 20.9 & 22.9 \\
& L2V-COT         & \textbf{41.8}& \textbf{25.5} & \textbf{26.9} \\
& w interpolation         & 31.1 & 16.3 & 19.4 \\
& w VLM       & 36.3 & 22.5 & 23.2 \\
\hline
\multirow{4}{*}{\textbf{InternVL}} 
& Direct Response         & 59.3 & 29.9 & 30.5  \\
& L2V-COT         & \textbf{61.6} & \textbf{33.3} & \textbf{33.7} \\
& w interpolation         & 53.6 & 21.3 & 19.1 \\
& w VLM         & 60.1 & 30.9 & 31.3 \\
\hline
\multirow{4}{*}{\textbf{QwenVL}} 
& Direct Response         & 60.5 & 26.9 & 33.8 \\
& L2V-COT         & \textbf{64.2} & \textbf{35.5} & \textbf{35.9} \\
& w interpolation         & 49.5 & 18.3 & 21.7 \\
& w VLM          & 61.4 & 28.4 & 33.9 \\
\hline
\end{tabular}
}
\vspace{-3pt}
\caption{Ablation study of L2V-CoT. "w interpolation" denotes replacing the LMN method with an interpolation-based method in L2V-COT.  “w VLM” uses VLM’s low-pass direction representation to activate its reasoning.}
\vspace{-10pt}
\label{table:ablation}
\end{table}

To demonstrate the effectiveness of the LPF method and LLM representations, we conduct an ablation study on L2V-CoT. Specifically, we test two variants: "w interpolation" replaces LPF with interpolation and "w VLM" uses low-pass direction representation for injection. The results are shown in Table~\ref{table:ablation}.
We find that interpolation leads to performance degradation. This is because interpolation causes significant loss of CoT information. As a result, it not only fails to improve the model’s reasoning ability, but also weakens its original performance.
Injecting VLM representations performs worse than injecting LLM representations. This is mainly because LLM representations help transfer the reasoning ability of the LLM, which is much stronger than that of the VLM.



\section{Conclusion}
In this paper, we empirically analyze CoT reasoning representations across modalities and architectures using LAT. Our analysis reveals that low-pass CoT direction representations effectively activate reasoning abilities in VLMs and exhibit similar distributions between VLMs and LLMs. Motivated by this insight, we propose L2V-CoT, a training-free method to transfer CoT reasoning capabilities from LLMs to VLMs. Specifically, we extract CoT direction representations from LLMs, performs frequency-domain resampling to align with VLM representation dimension, and injects the resampled representation into VLMs. Extensive experiments demonstrate that L2V-CoT significantly enhances VLM reasoning performance, achieving an average improvement of 3.7\% and up to 8.6\% across various benchmarks.

\section{Acknowledgments}
The work is supported by the National Natural Science Foundation of China (No. 92270118 and No. 62276269) and the Beijing Natural Science Foundation (No. 1232009). 


\bibliography{aaai2026}

@article{xia2024beyond,
  title={Beyond chain-of-thought: A survey of chain-of-x paradigms for llms},
  author={Xia, Yu and Wang, Rui and Liu, Xu and Li, Mingyan and Yu, Tong and Chen, Xiang and McAuley, Julian and Li, Shuai},
  journal={arXiv preprint arXiv:2404.15676},
  year={2024}
}

@inproceedings{zhang2025enhancing,
  title={Enhancing chain of thought prompting in large language models via reasoning patterns},
  author={Zhang, Yufeng and Wang, Xuepeng and Wu, Lingxiang and Wang, Jinqiao},
  booktitle={Proceedings of the AAAI Conference on Artificial Intelligence},
  volume={39},
  number={24},
  pages={25985--25993},
  year={2025}
}

@inproceedings{cheng2025comt,
  title={Comt: A novel benchmark for chain of multi-modal thought on large vision-language models},
  author={Cheng, Zihui and Chen, Qiguang and Zhang, Jin and Fei, Hao and Feng, Xiaocheng and Che, Wanxiang and Li, Min and Qin, Libo},
  booktitle={Proceedings of the AAAI Conference on Artificial Intelligence},
  volume={39},
  number={22},
  pages={23678--23686},
  year={2025}
}

@article{chen2025bring,
  title={Bring reason to vision: Understanding perception and reasoning through model merging},
  author={Chen, Shiqi and Zhang, Jinghan and Zhu, Tongyao and Liu, Wei and Gao, Siyang and Xiong, Miao and Li, Manling and He, Junxian},
  journal={arXiv preprint arXiv:2505.05464},
  year={2025}
}

@inproceedings{zhang2024mathverse,
  title={Mathverse: Does your multi-modal llm truly see the diagrams in visual math problems?},
  author={Zhang, Renrui and Jiang, Dongzhi and Zhang, Yichi and Lin, Haokun and Guo, Ziyu and Qiu, Pengshuo and Zhou, Aojun and Lu, Pan and Chang, Kai-Wei and Qiao, Yu and others},
  booktitle={European Conference on Computer Vision},
  pages={169--186},
  year={2024},
  organization={Springer}
}

@article{du2025virgo,
  title={Virgo: A preliminary exploration on reproducing o1-like mllm},
  author={Du, Yifan and Liu, Zikang and Li, Yifan and Zhao, Wayne Xin and Huo, Yuqi and Wang, Bingning and Chen, Weipeng and Liu, Zheng and Wang, Zhongyuan and Wen, Ji-Rong},
  journal={arXiv preprint arXiv:2501.01904},
  year={2025}
}

@article{zou2023representation,
  title={Representation engineering: A top-down approach to ai transparency},
  author={Zou, Andy and Phan, Long and Chen, Sarah and Campbell, James and Guo, Phillip and Ren, Richard and Pan, Alexander and Yin, Xuwang and Mazeika, Mantas and Dombrowski, Ann-Kathrin and others},
  journal={arXiv preprint arXiv:2310.01405},
  year={2023}
}

@article{tang2025unlocking,
  title={Unlocking General Long Chain-of-Thought Reasoning Capabilities of Large Language Models via Representation Engineering},
  author={Tang, Xinyu and Wang, Xiaolei and Lv, Zhihao and Min, Yingqian and Zhao, Wayne Xin and Hu, Binbin and Liu, Ziqi and Zhang, Zhiqiang},
  journal={CoRR},
  year={2025}
}

@article{yangadapting,
  title={Adapting multi-modal large language model to concept drift from pre-training onwards},
  author={Yang, Xiaoyu and Lu, Jie and Yu, En},
  journal={arXiv preprint arXiv:2405.13459},
  year={2024}
}

@inproceedings{chen-etal-2024-m3cot,
    title = "{M}$^3${C}o{T}: A Novel Benchmark for Multi-Domain Multi-step Multi-modal Chain-of-Thought",
    author = "Chen, Qiguang  and
      Qin, Libo  and
      Zhang, Jin  and
      Chen, Zhi  and
      Xu, Xiao  and
      Che, Wanxiang",
    editor = "Ku, Lun-Wei  and
      Martins, Andre  and
      Srikumar, Vivek",
    booktitle = "Proceedings of the 62nd Annual Meeting of the Association for Computational Linguistics (Volume 1: Long Papers)",
    month = aug,
    year = "2024",
    address = "Bangkok, Thailand",
    publisher = "Association for Computational Linguistics",
    url = "https://aclanthology.org/2024.acl-long.446/",
    doi = "10.18653/v1/2024.acl-long.446",
    pages = "8199--8221"
}

@article{thawakar2025llamav,
  title={Llamav-o1: Rethinking step-by-step visual reasoning in llms},
  author={Thawakar, Omkar and Dissanayake, Dinura and More, Ketan and Thawkar, Ritesh and Heakl, Ahmed and Ahsan, Noor and Li, Yuhao and Zumri, Mohammed and Lahoud, Jean and Anwer, Rao Muhammad and others},
  journal={arXiv preprint arXiv:2501.06186},
  year={2025}
}

@article{wu2025boosting,
  title={Boosting multimodal reasoning with mcts-automated structured thinking},
  author={Wu, Jinyang and Feng, Mingkuan and Zhang, Shuai and Jin, Ruihan and Che, Feihu and Wen, Zengqi and Tao, Jianhua},
  journal={arXiv e-prints},
  pages={arXiv--2502},
  year={2025}
}

@article{yao2024mulberry,
  title={Mulberry: Empowering mllm with o1-like reasoning and reflection via collective monte carlo tree search},
  author={Yao, Huanjin and Huang, Jiaxing and Wu, Wenhao and Zhang, Jingyi and Wang, Yibo and Liu, Shunyu and Wang, Yingjie and Song, Yuxin and Feng, Haocheng and Shen, Li and others},
  journal={arXiv preprint arXiv:2412.18319},
  year={2024}
}

@article{luo2025ursa,
  title={Ursa: Understanding and verifying chain-of-thought reasoning in multimodal mathematics},
  author={Luo, Ruilin and Zheng, Zhuofan and Wang, Yifan and Ni, Xinzhe and Lin, Zicheng and Jiang, Songtao and Yu, Yiyao and Shi, Chufan and Chu, Ruihang and Zeng, Jin and others},
  journal={arXiv preprint arXiv:2501.04686},
  year={2025}
}

@article{zhang2023multimodal,
  title={Multimodal chain-of-thought reasoning in language models},
  author={Zhang, Zhuosheng and Zhang, Aston and Li, Mu and Zhao, Hai and Karypis, George and Smola, Alex},
  journal={arXiv preprint arXiv:2302.00923},
  year={2023}
}

@article{li2023inference,
  title={Inference-time intervention: Eliciting truthful answers from a language model},
  author={Li, Kenneth and Patel, Oam and Vi{\'e}gas, Fernanda and Pfister, Hanspeter and Wattenberg, Martin},
  journal={Advances in Neural Information Processing Systems},
  volume={36},
  pages={41451--41530},
  year={2023}
}

@inproceedings{hollinsworth-etal-2024-language,
    title = "Language Models Linearly Represent Sentiment",
    author = "Tigges, Curt  and
      Hollinsworth, Oskar J.  and
      Geiger, Atticus  and
      Nanda, Neel",
    editor = "Belinkov, Yonatan  and
      Kim, Najoung  and
      Jumelet, Jaap  and
      Mohebbi, Hosein  and
      Mueller, Aaron  and
      Chen, Hanjie",
    booktitle = "Proceedings of the 7th BlackboxNLP Workshop: Analyzing and Interpreting Neural Networks for NLP",
    month = nov,
    year = "2024",
    address = "Miami, Florida, US",
    publisher = "Association for Computational Linguistics",
    url = "https://aclanthology.org/2024.blackboxnlp-1.5/",
    doi = "10.18653/v1/2024.blackboxnlp-1.5",
    pages = "58--87"
}

@inproceedings{kim2024structure,
  title={Structure-aware multimodal sequential learning for visual dialog},
  author={Kim, Young-Jin and Kim, Min-Jun and An, Kyunghwan and Ahn, Jinwoo and Kim, Jaeseok and Heo, Yu-Jung and Chang, Du-Seong and Kim, Eun-Sol},
  booktitle={Proceedings of the AAAI Conference on Artificial Intelligence},
  volume={38},
  number={12},
  pages={13193--13201},
  year={2024}
}

@misc{openai2024learningtoreason,
  title        = {Learning to Reason with Large Language Models},
  author       = {{OpenAI}},
  howpublished = {OpenAI Blog and arXiv preprint},
  year         = {2024},
  note         = {\url{https://openai.com/index/learning-to-reason-with-llms/} (accessed 2024-09-29)},
}

@article{guo2025deepseek,
  title={Deepseek-r1: Incentivizing reasoning capability in llms via reinforcement learning},
  author={Guo, Daya and Yang, Dejian and Zhang, Haowei and Song, Junxiao and Zhang, Ruoyu and Xu, Runxin and Zhu, Qihao and Ma, Shirong and Wang, Peiyi and Bi, Xiao and others},
  journal={arXiv preprint arXiv:2501.12948},
  year={2025}
}

@article{jolliffe2002springer,
  title={Springer series in statistics},
  author={Jolliffe, Ian T},
  journal={Principal component analysis},
  volume={29},
  pages={912},
  year={2002},
  publisher={Springer-Verlag}
}

@article{meta2024introducing,
  title={Introducing meta llama 3: The most capable openly available llm to date},
  author={Meta, AI},
  journal={Meta AI},
  volume={2},
  number={5},
  pages={6},
  year={2024}
}

@article{burns2022discovering,
  title={Discovering latent knowledge in language models without supervision},
  author={Burns, Collin and Ye, Haotian and Klein, Dan and Steinhardt, Jacob},
  journal={arXiv preprint arXiv:2212.03827},
  year={2022}
}

@article{liu2025we,
  title={Do we Really Need Visual Instructions? Towards Visual Instruction-Free Fine-tuning for Large Vision-Language Models},
  author={Liu, Zikang and Zhou, Kun and Zhao, Wayne Xin and Gao, Dawei and Li, Yaliang and Wen, Ji-Rong},
  journal={arXiv preprint arXiv:2502.11427},
  year={2025}
}

@article{min2024imitate,
  title={Imitate, explore, and self-improve: A reproduction report on slow-thinking reasoning systems},
  author={Min, Yingqian and Chen, Zhipeng and Jiang, Jinhao and Chen, Jie and Deng, Jia and Hu, Yiwen and Tang, Yiru and Wang, Jiapeng and Cheng, Xiaoxue and Song, Huatong and others},
  journal={arXiv preprint arXiv:2412.09413},
  year={2024}
}

@article{lu2023mathvista,
  title={Mathvista: Evaluating mathematical reasoning of foundation models in visual contexts},
  author={Lu, Pan and Bansal, Hritik and Xia, Tony and Liu, Jiacheng and Li, Chunyuan and Hajishirzi, Hannaneh and Cheng, Hao and Chang, Kai-Wei and Galley, Michel and Gao, Jianfeng},
  journal={arXiv preprint arXiv:2310.02255},
  year={2023}
}

@article{wang2024measuring,
  title={Measuring multimodal mathematical reasoning with math-vision dataset},
  author={Wang, Ke and Pan, Junting and Shi, Weikang and Lu, Zimu and Ren, Houxing and Zhou, Aojun and Zhan, Mingjie and Li, Hongsheng},
  journal={Advances in Neural Information Processing Systems},
  volume={37},
  pages={95095--95169},
  year={2024}
}

@article{zou2024dynamath,
  title={Dynamath: A dynamic visual benchmark for evaluating mathematical reasoning robustness of vision language models},
  author={Zou, Chengke and Guo, Xingang and Yang, Rui and Zhang, Junyu and Hu, Bin and Zhang, Huan},
  journal={arXiv preprint arXiv:2411.00836},
  year={2024}
}

@article{chen2024we,
  title={Are we on the right way for evaluating large vision-language models?},
  author={Chen, Lin and Li, Jinsong and Dong, Xiaoyi and Zhang, Pan and Zang, Yuhang and Chen, Zehui and Duan, Haodong and Wang, Jiaqi and Qiao, Yu and Lin, Dahua and others},
  journal={Advances in Neural Information Processing Systems},
  volume={37},
  pages={27056--27087},
  year={2024}
}

@article{hu2024understanding,
  title={Understanding reasoning in chain-of-thought from the hopfieldian view},
  author={Hu, Lijie and Liu, Liang and Yang, Shu and Chen, Xin and Tan, Zhen and Ali, Muhammad Asif and Li, Mengdi and Wang, Di},
  journal={arXiv preprint arXiv:2410.03595},
  year={2024}
}

@article{christ2024math,
  title={Math Neurosurgery: Isolating Language Models' Math Reasoning Abilities Using Only Forward Passes},
  author={Christ, Bryan R and Gottesman, Zack and Kropko, Jonathan and Hartvigsen, Thomas},
  journal={arXiv preprint arXiv:2410.16930},
  year={2024}
}

@article{liu2024llava,
  title={Llava-next: Improved reasoning, ocr, and world knowledge, January 2024},
  author={Liu, Haotian and Li, Chunyuan and Li, Yuheng and Li, Bo and Zhang, Yuanhan and Shen, Sheng and Lee, Yong Jae},
  journal={URL https://llava-vl. github. io/blog/2024-01-30-llava-next},
  volume={1},
  number={8},
  year={2024}
}

@article{laurenccon2024matters,
  title={What matters when building vision-language models?},
  author={Lauren{\c{c}}on, Hugo and Tronchon, L{\'e}o and Cord, Matthieu and Sanh, Victor},
  journal={Advances in Neural Information Processing Systems},
  volume={37},
  pages={87874--87907},
  year={2024}
}

@inproceedings{chen2024internvl,
  title={Internvl: Scaling up vision foundation models and aligning for generic visual-linguistic tasks},
  author={Chen, Zhe and Wu, Jiannan and Wang, Wenhai and Su, Weijie and Chen, Guo and Xing, Sen and Zhong, Muyan and Zhang, Qinglong and Zhu, Xizhou and Lu, Lewei and others},
  booktitle={Proceedings of the IEEE/CVF conference on computer vision and pattern recognition},
  pages={24185--24198},
  year={2024}
}

@article{wang2024qwen2,
  title={Qwen2-vl: Enhancing vision-language model's perception of the world at any resolution},
  author={Wang, Peng and Bai, Shuai and Tan, Sinan and Wang, Shijie and Fan, Zhihao and Bai, Jinze and Chen, Keqin and Liu, Xuejing and Wang, Jialin and Ge, Wenbin and others},
  journal={arXiv preprint arXiv:2409.12191},
  year={2024}
}

@article{hartsock2024vision,
  title={Vision-language models for medical report generation and visual question answering: A review},
  author={Hartsock, Iryna and Rasool, Ghulam},
  journal={Frontiers in artificial intelligence},
  volume={7},
  pages={1430984},
  year={2024},
  publisher={Frontiers Media SA}
}

@article{gerlach2024waveform,
  title={Waveform resampling with LMN method},
  author={Gerlach, Lino and Gu, Wenqiang and Nayak, Nitish and Qian, Xin and Viren, Brett},
  journal={Journal of Instrumentation},
  volume={19},
  number={10},
  pages={P10029},
  year={2024},
  publisher={IOP Publishing}
}

@article{kelkar2007extension,
  title={An extension of Parseval's theorem and its use in calculating transient energy in the frequency domain},
  author={Kelkar, SS and Grigsby, LL and Langsner, J},
  journal={IEEE Transactions on Industrial Electronics},
  number={1},
  pages={42--45},
  year={2007},
  publisher={IEEE}
}

@inproceedings{parkgeneralizing,
  title={Generalizing from SIMPLE to HARD Visual Reasoning: Can We Mitigate Modality Imbalance in VLMs?},
  author={Park, Simon and Panigrahi, Abhishek and Cheng, Yun and Yu, Dingli and Goyal, Anirudh and Arora, Sanjeev},
  booktitle={Forty-second International Conference on Machine Learning}
}

@inproceedings{schroditwo,
  title={Two Effects, One Trigger: On the Modality Gap, Object Bias, and Information Imbalance in Contrastive Vision-Language Models},
  author={Schrodi, Simon and Hoffmann, David T and Argus, Max and Fischer, Volker and Brox, Thomas},
  booktitle={The Thirteenth International Conference on Learning Representations}
}

@article{tang2025enhancing,
  title={Enhancing Cross-task Transfer of Large Language Models via Activation Steering},
  author={Tang, Xinyu and Lv, Zhihao and Cheng, Xiaoxue and Li, Junyi and Zhao, Wayne Xin and Wen, Zujie and Zhang, Zhiqiang and Zhou, Jun},
  journal={arXiv preprint arXiv:2507.13236},
  year={2025}
}

@article{wang2025bee,
  title={BEE-RAG: Balanced Entropy Engineering for Retrieval-Augmented Generation},
  author={Wang, Yuhao and Ren, Ruiyang and Wang, Yucheng and Liu, Jing and Zhao, Wayne Xin and Wu, Hua and Wang, Haifeng},
  journal={arXiv preprint arXiv:2508.05100},
  year={2025}
}

@inproceedings{wang2025unveiling,
  title={Unveiling Knowledge Utilization Mechanisms in LLM-based Retrieval-Augmented Generation},
  author={Wang, Yuhao and Ren, Ruiyang and Wang, Yucheng and Zhao, Wayne Xin and Liu, Jing and Wu, Hua and Wang, Haifeng},
  booktitle={Proceedings of the 48th International ACM SIGIR Conference on Research and Development in Information Retrieval},
  pages={1262--1271},
  year={2025}
}

@inproceedings{li2024images,
  title={Images are Achilles’ heel of alignment: Exploiting visual vulnerabilities for jailbreaking multimodal large language models},
  author={Li, Yifan and Guo, Hangyu and Zhou, Kun and Zhao, Wayne Xin and Wen, Ji-Rong},
  booktitle={European Conference on Computer Vision},
  pages={174--189},
  year={2024},
  organization={Springer}
}

@article{li2025analyzing,
  title={Analyzing and Mitigating Object Hallucination: A Training Bias Perspective},
  author={Li, Yifan and Zhou, Kun and Zhao, Wayne Xin and Fang, Lei and Wen, Ji-Rong},
  journal={arXiv preprint arXiv:2508.04567},
  year={2025}
}

@article{li2025unleashing,
  title={Unleashing Perception-Time Scaling to Multimodal Reasoning Models},
  author={Li, Yifan and Chen, Zhenghao and Wu, Ziheng and Zhou, Kun and Luo, Ruipu and Zhang, Can and He, Zhentao and Zhan, Yufei and Zhao, Wayne Xin and Qiu, Minghui},
  journal={arXiv preprint arXiv:2510.08964},
  year={2025}
}

@article{zhan2024over,
  title={Over-parameterized student model via tensor decomposition boosted knowledge distillation},
  author={Zhan, Yu-Liang and Lu, Zhong-Yi and Sun, Hao and Gao, Ze-Feng},
  journal={Advances in Neural Information Processing Systems},
  volume={37},
  pages={69445--69470},
  year={2024}
}
\clearpage
\newpage

\include{appendix}

\setcounter{secnumdepth}{2}

\setcounter{figure}{0}
\setcounter{table}{0}
\setcounter{page}{1}
\setcounter{equation}{0}
\setcounter{algorithm}{0}
\setcounter{definition}{0}

\renewcommand{\theequation}{S.\arabic{equation}}
\renewcommand{\thefigure}{S.\arabic{figure}}
\renewcommand{\thetable}{S.\arabic{table}}
\renewcommand{\thealgorithm}{S.\arabic{algorithm}}
\appendix
\begin{center}
{\centering \Large \textbf{APPENDIX}}
\end{center}

\section{Detailed Description of Baselines}
\label{sec:appendix_Baseline}

\citet{tang2025unlocking} selected representative baselines to demonstrate the effectiveness of the proposed GLoRE method when showing that activation engineering can unlock the long CoT abilities of LLMs. We follow GLoRE and use baselines: Non-CoT response, Few-shot CoT, MathNeuro~\cite{christ2024math}, and RoT~\cite{hu2024understanding}. In addition, we include several VLM-specific methods that aim to transfer reasoning from LLMs as baselines: Finetuned CoT~\cite{du2025virgo} and Model Merging~\cite{chen2025bring}. In this section, we describe each baseline in detail.

\textbf{Non-CoT response:} This method uses a vanilla VLM prompted with a CoT-style instruction (e.g., "Answer the question step by step and put the final answer in \textbackslash\textbackslash boxed\{\}"). The VLM responds to the question and places the final answer inside \textbackslash \textbackslash boxed\{\}.

\textbf{Few-shot CoT:} This method provides the VLM with CoT answers generated by an LLM as examples. The model learns to imitate the reasoning ability of LLM through instruction-based prompting.

\textbf{MathNeuro}~\cite{christ2024math}:By analyzing the forward pass weights and activations, MathNeuro pinpoints parameters crucial for reasoning and selectively refines them via pruning and scaling.

\textbf{RoT}~\cite{hu2024understanding}: This method also uses representation injection. However, unlike our approach which injects representations from CoT answers, RoT extracts representations from CoT prompts and injects them into the VLM.

\textbf{Finetuned CoT}~\cite{du2025virgo}: This method directly fine-tunes the VLM using the same data we use to obtain CoT direction representations. The goal is to teach the model the CoT reasoning pattern through supervised learning.

\textbf{Model Merging}~\cite{chen2025bring}: This method merges a VLM and an LLM that share the same backbone. It transfers the reasoning ability from the LLM to the VLM to enhance the model’s reasoning performance.
\section{Spectral Similarity Between VLMs and LLMs}
\label{sec:appendix_energy}
\begin{figure}[t!]
    \centering
    \small
    \includegraphics[width=\linewidth]{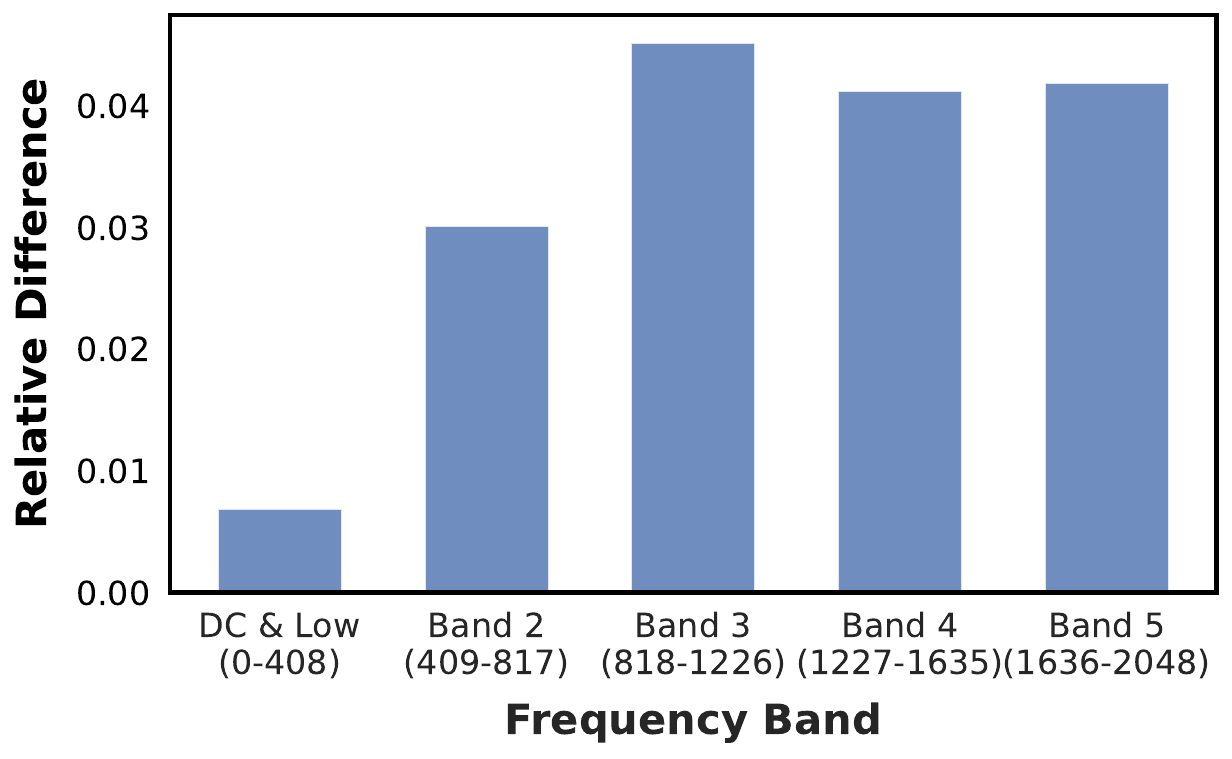}
    %
    \caption{Relative energy difference across frequency bands between the CoT direction representations of the VLM (Qwen2-VL-7B-Instruct) and the LLM (LLaMA3-8B). "DC\&Low" denotes the low-frequency band, while the others correspond to mid- and high-frequency bands.}
    \label{fig:energy}
\end{figure}

In signal processing, differences between signals are often evaluated by comparing their energy across frequency bands~\cite{kelkar2007extension}. We compute the relative error of CoT direction representations between LLM and VLM in each frequency band. As shown in Figure~\ref{fig:energy}, the error in the low-frequency band (Band 1) is much smaller than in the high-frequency bands. This further supports our empirical finding: the low-pass filtered direction representations of VLM are close to those of LLM.

\section{Stronger LLMs Enable Stronger Performance Transfer.}
We further investigate how the capability of the LLM influences the transferred performance of the VLM. Specifically, we use different sizes of DeepSeek-R1-Distill-Qwen to extract CoT pattern representations. As shown in Table~\ref{table:diffllm}, L2M-CoT consistently improves performance across all models and datasets, with stronger LLMs yielding greater gains in VLM reasoning ability after transfer.
\begin{table}[t!]
\centering
\scalebox{0.78}{
\begin{tabular}{c|c|c|c|c}
\hline
\textbf{Model} & \textbf{Method} & \textbf{MathVista} & \textbf{MathVerse} & \textbf{DynaMath} \\
\hline
\multirow{4}{*}{\textbf{LLAVA}} 

& Direct Response         & 35.2 & 20.9 & 22.9 \\
& L2V-COT-7B         & 38.6 & 23.7 & 24.2 \\
& L2V-COT-14B         & 38.9 & 24.4 & 24.9 \\
& L2V-COT-32B       & \textbf{41.8}& \textbf{25.5} & \textbf{26.9} \\
\hline
\multirow{4}{*}{\textbf{InternVL}} 
& Direct Response         & 59.3 & 29.9 & 30.5  \\
& L2V-COT-7B         & 60.9 & 31.1 & 31.6 \\
& L2V-COT-14B        & 61.0 & 31.9 & 32.6 \\
& L2V-COT-32B        & \textbf{61.6} & \textbf{33.3} & \textbf{33.7} \\
\hline
\multirow{4}{*}{\textbf{QwenVL}} 
& Direct Response         & 60.5 & 26.9 & 33.8 \\
& L2V-COT-7B         & 61.6 & 29.4 & 34.2 \\
& L2V-COT-14B         & 62.3 & 31.8 & 34.5 \\
& L2V-COT-32B         & \textbf{64.2} & \textbf{35.5} & \textbf{35.9} \\
\hline
\end{tabular}
}
\caption{Performance of L2V-CoT with different LLM representation. 7B, 14B, and 32B refer to the model sizes of DeepSeek-R1-Distill-Qwen.}
\vspace{-15pt}
\label{table:diffllm}
\end{table}
\begin{table*}[t!]
\centering
\scalebox{1}{
\begin{tabular}{c|ccc|cccccc|ccc|c|c}
\hline
\multirow{2}{*}{\textbf{Model}}
  & \multicolumn{3}{c|}{\textbf{MathVista}}
  & \multicolumn{6}{c|}{\textbf{MathVerse Benchmarks}}
  & \multicolumn{3}{c|}{\textbf{MMStar}}
  & \multirow{2}{*}{\textbf{DM}}
  & \multirow{2}{*}{\textbf{MV}} \\
\cline{2-13}
  & All & General & Math 
  & Overall & T-D & T-L & V-I & V-D & V-O 
  & All & Percep. & Reason. 
  &  &  \\
\hline
LLAVA    & 12 & 12 & 12 & 17 & 16 & 13 & 14 & 18 & 19 & 15 & 15 & 15 & 17 & 16 \\
InternVL & 19 & 19 & 19 & 16 & 17 & 14 & 19 & 13 & 20 & 17 & 17 & 17 & 18 & 15 \\
QwenVL   & 9 & 9 & 9 & 16 & 17 & 12 & 13 & 14 & 16 & 12 & 12 & 12 & 15 & 14 \\
Idefics  & 15 & 15 & 15 & 20 & 19 & 16 & 14 & 21 & 12 & 13 & 13 & 13 & 16 & 17 \\
\hline
\end{tabular}
}
\caption{Injection layer of the VLM.}
\label{table:layer}
\end{table*}

\begin{table*}[t!]
\centering
\scalebox{1}{
\begin{tabular}{c|ccc|cccccc|ccc|c|c}
\hline
\multirow{2}{*}{\textbf{Model}}
  & \multicolumn{3}{c|}{\textbf{MathVista}}
  & \multicolumn{6}{c|}{\textbf{MathVerse Benchmarks}}
  & \multicolumn{3}{c|}{\textbf{MMStar}}
  & \multirow{2}{*}{\textbf{DM}}
  & \multirow{2}{*}{\textbf{MV}} \\
\cline{2-13}
  & All & General & Math 
  & Overall & T-D & T-L & V-I & V-D & V-O 
  & All & Percep. & Reason. 
  &  &  \\
\hline
LLAVA    & 0.3 & 0.3 & 0.3 & 0.6 & 0.5 & 0.4 & 0.6 & 0.5 & 0.6 & 0.3 & 0.3 & 0.3 & 0.5 & 0.4 \\
InternVL & 0.4 & 0.4 & 0.3 & 0.7 & 0.6 & 0.3 & 0.7 & 0.4 & 0.6 & 0.6 & 0.6 & 0.6 & 0.5 & 0.6 \\
QwenVL   & 0.3 & 0.3 & 0.3 & 0.5 & 0.5 & 0.3 & 0.4 & 0.5 & 0.6 & 0.5 & 0.5 & 0.5 & 0.4 & 0.4 \\
Idefics  & 0.5 & 0.5 & 0.5 & 0.8 & 0.6 & 0.5 & 0.4 & 0.7 & 0.3 & 0.6 & 0.6 & 0.6 & 0.5 & 0.5 \\
\hline
\end{tabular}
}
\caption{Injection strength.}
\label{table:strength}
\end{table*}

\section{L2V-CoT Algorithm.}
\floatname{algorithm}{Algorithm}
\begin{algorithm*}[t!]
\small
    \caption{L2V-CoT for transfering the reasoning apability of LLM to VLM.} 
    \begin{algorithmic}[1] 
        \Require A set of reasoning problems $\{q_i\}_{i=1}^n$, a reasoning-capable LLM $M_{\text{LLM}}$, a vanilla VLM $M_{\text{VLM}}$, A pair of question transfer template $\mathcal{T}^+_{CoT}$ and $\mathcal{T}^-_{CoT}$.
        \State Use $\mathcal{T}^+{\text{CoT}}$ and $\mathcal{T}^-{\text{CoT}}$ to convert ${q_i}_{i=1}^n$ into question sets $\{q^+_i\}_{i=1}^n$ and $\{q^-_i\}_{i=1}^n$
        \State Use $\{q^+_i\}_{i=1}^n$ and $\{q^-_i\}_{i=1}^n$ to obtain CoT responses $\{c_i\}_{i=1}^n$ and direction responses $\{d_i\}_{i=1}^n$ from $M_{\text{LLM}}$
        \State Use $\{c_i\}_{i=1}^n$ and $\{d_i\}_{i=1}^n$ to extract CoT representations $h_L(c_i,l_L)$ and non-CoT representations $h_L(d_i,l_L)$ from the $M_{\text{LLM}}$
        \State Construct CoT direction representations $\{u_i(l_L)\}_{i=1}^n=\{h_L(c_i,l_L)-h_L(d_i,l_L)\}_{i=1}^n$
        \State $v(l_L) \gets Mean(\{u_i(l_L)\}_{i=1}^n)$
        \State $v_{LPF}(l_L) \gets LPF(V(l_L))$
        \State $\hat{v}_{LPF}(l_L) \gets {v}_{LPF}(l_L)$
        \State $h_V(l_V) \gets \hat{v}_{LPF}(l_L)$ + $h_V(l_V)$
        \State $\hat{h}_V(l_V) \gets Normalize({h}_V(l_V))$
        \State replace ${h}_V(l_V)$ in $M_{\text{VLM}}$ with $\hat{h}_V(l_V)$
        \end{algorithmic}
\label{alg:L2V_CoT}
\end{algorithm*}

We propose L2V-CoT to transfer the reasoning ability of the LLM to the VLM, thereby enhancing the reasoning capability of the VLM. The detailed algorithm is shown in Algorithm~\ref{alg:L2V_CoT}.
\section{Experimental Details}
\label{sec:appendix_experiment_detail}
In our experiments, we inject the representations from the LLM into the VLM. Therefore, the extraction layer of the LLM, the injection strength, and the injection layer of the VLM are key hyperparameters. We divide the layers of both the LLM and VLM into three categories: low, middle, and high. Then we inject the middle-layer representations from the LLM into the VLM. The specific injection layer and strength for the VLM are shown in Table~\ref{table:layer} and Table~\ref{table:strength}. The tasks in MathVista and MMStar share the same hyperparameters because they belong to the same dataset. Therefore, we use the same hyperparameters for training. In contrast, each task in MathVerse has a different training set, so we use different hyperparameters accordingly.

We show all the checkpoints we use for experiments in Table~\ref{table:checkpoint}.
\begin{table}[t!]
\centering
\scalebox{0.7}{
\begin{tabular}{c|c}
\hline
\textbf{Model} & \textbf{Huggingface Checkpoint} \\
\hline
\textbf{LLaVA-Next-LLaMA3-8B}     & llava-hf/llama3-llava-next-8b-hf \\
\textbf{InternVL2-8B}  & OpenGVLab/InternVL2-8B \\
\textbf{Qwen2-VL-7B-Instruc}    & Qwen/Qwen2-VL-7B-Instruct \\
\textbf{Idefics2-8B}    & HuggingFaceM4/idefics2-8b \\
\textbf{InternVL2-26B}    & HuggingFaceM4/idefics2-26b \\
\textbf{InternVL2-40B}    & HuggingFaceM4/idefics2-40b \\
\textbf{DeepSeek-R1-Distill-Qwen-32} & deepseek-ai/DeepSeek-R1-Distill-Qwen-32B \\

\hline
\end{tabular}
}
\caption{All the Huggingface checkpoints used in our experiments.}

\label{table:checkpoint}
\end{table}

\section{Dataset Details}
\label{sec:appendix_dataset}
\textbf{MathVista}~\cite{lu2023mathvista}:
This dataset is designed to evaluate the performance of vision-language models on multimodal math problem solving. We conduct our experiments on MathVista MINI, a subset of the full MathVista dataset. MathVista MINI is mainly used for quick experiments or training under limited resources. It contains around 1,000 questions selected from the full MathVista dataset. The questions are divided into two categories: General and Math.

\textbf{MathVerse}~\cite{zhang2024mathverse}:
MathVerse is a large-scale benchmark for multimodal mathematical reasoning. It consists of multiple datasets from different sources, including MathQA+Draw, GSM8K+Diagrams, and SVAMP+Tables. The tasks cover various types of reasoning, such as arithmetic, algebraic, and logical reasoning, all paired with visual elements. The goal is to provide a unified evaluation of multimodal LLMs on different types of mathematical problems. Each subset can be used as an individual task dataset, which helps analyze the challenges of different question types. The tasks are categorized into six types: V-I (vision intensive), V-D (vision dominant), V-O (vision only), T-D (text dominant), T-L (text lite), and T-O (text only).

\textbf{MMStar}~\cite{chen2024we}:
MMStar is a comprehensive multimodal math dataset. Its questions are usually closer to real-life scenarios and have longer descriptions. The dataset requires models to have basic commonsense reasoning and strong language understanding. The tasks are divided into perception tasks and reasoning tasks.

Dynamath~\cite{zou2024dynamath}:
It is a dataset designed to evaluate multimodal dynamic math reasoning ability. It combines visual understanding, temporal reasoning, and mathematical modeling. The model needs to perform image modeling across time and demonstrate reasoning and memory abilities.

\textbf{MathVision}~\cite{wang2024measuring}:
MathVision is a multimodal reasoning dataset focused on visual math problems. It evaluates the model’s ability to combine image understanding with mathematical reasoning. The model needs to understand text, recognize objects in images, read chart data, and interpret visual relationships.

\section{Experiment Platform}
\label{Platform}
We conduct comprehensive experiments utilizing the NVIDIA H800 80GB PCIe paired with the Intel(R) Xeon(R) Platinum 8380 CPU @ 2.30GHz. 

\section{Case Study}
\label{sec:appendix_case}
\begin{figure*}[t!]
    \centering
    \small
    \includegraphics[width=\linewidth]{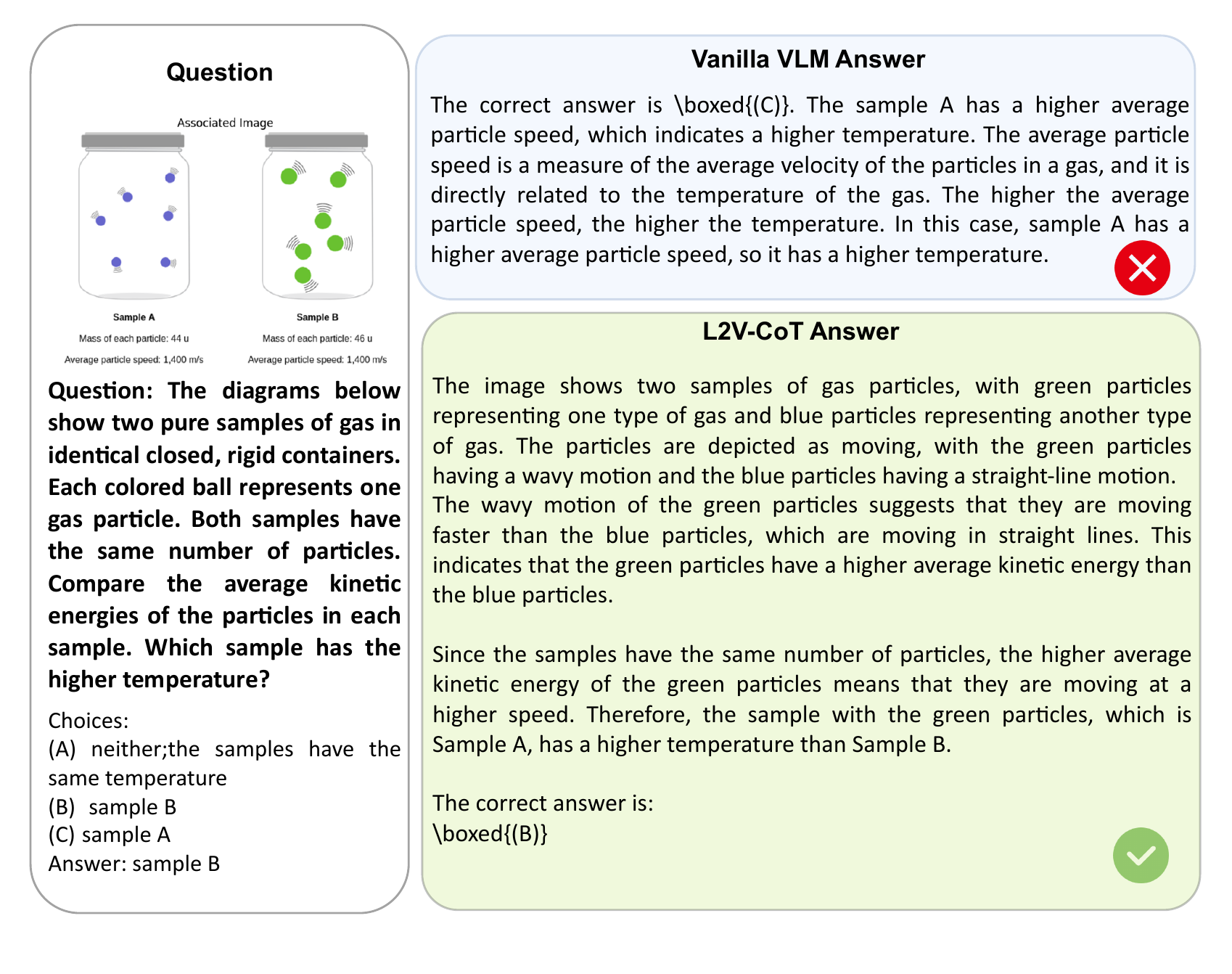}
    \caption{An example of L2V-CoT transferring reasoning ability from LLM to VLM. Vanilla VLMs first give the answer and then provide an explanation. In contrast, L2V-CoT reasons for a longer time before producing the answer.}
    \label{fig:case1}
\end{figure*}

\begin{figure*}[t!]
    \centering
    \small
    \includegraphics[width=\linewidth]{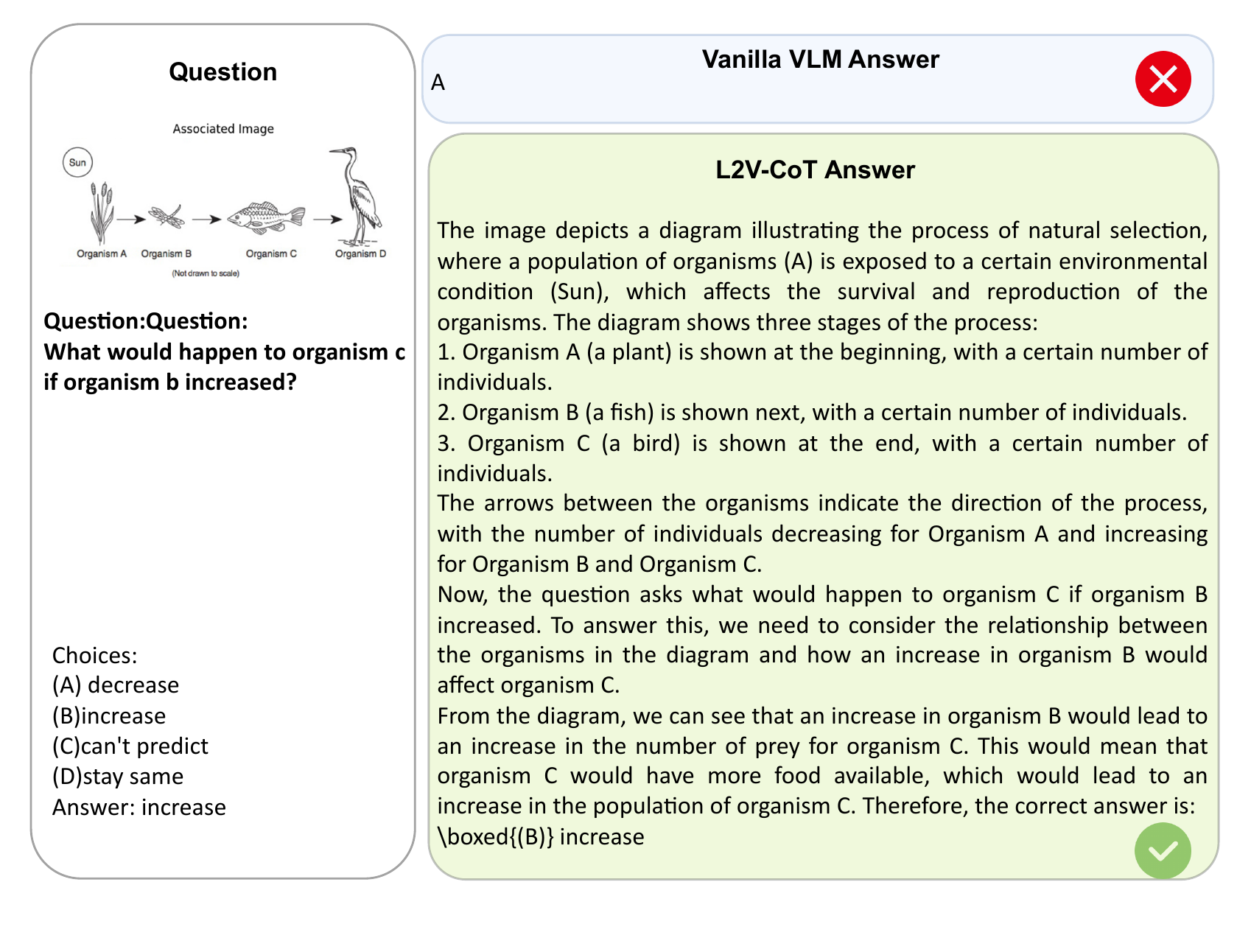}
    \caption{An example of L2V-CoT transferring reasoning ability from LLM to VLM. The vanilla VLM tends to answer immediately. L2V-CoT, however, answers after performing extended reasoning.}
    \label{fig:case2}
\end{figure*}

\begin{figure*}[t!]
    \centering
    \small
    \includegraphics[width=\linewidth]{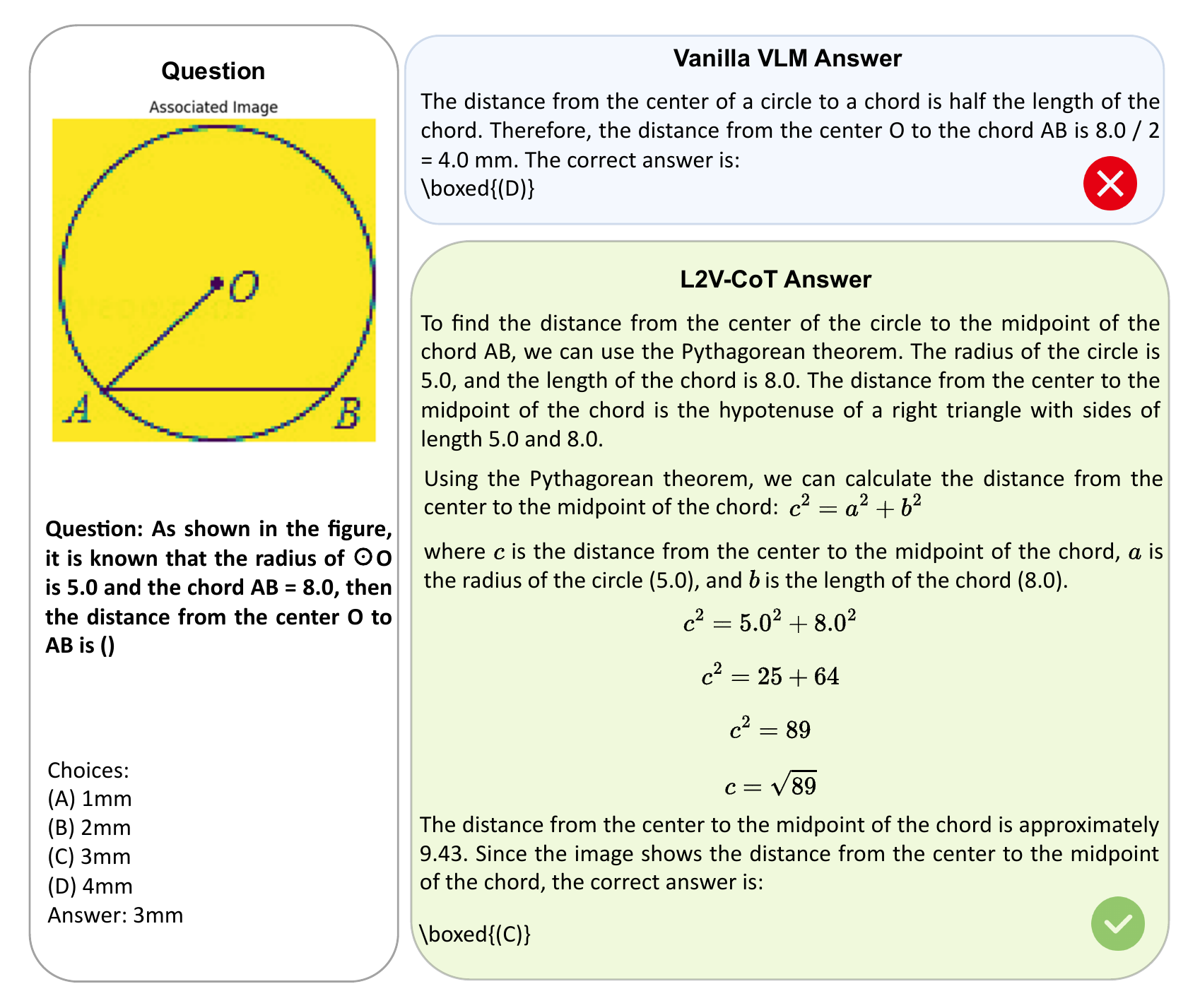}
    \caption{An example of L2V-CoT transferring reasoning ability from LLM to VLM. It can be seen that vanilla VLMs perform short reasoning. In contrast, L2V-CoT generates the answer after a longer reasoning process.}
    \label{fig:case3}
\end{figure*}

In this section, we present several examples from n LLaVA-Next-LLaMA3-8B to intuitively demonstrate that our method can enhance the reasoning capability of VLMs. The case study examples are show in Figure~\ref{fig:case1}, Figure~\ref{fig:case2} and Figure~\ref{fig:case3}. We can observe that our method indeed improves the reasoning ability of VLMs.

\end{document}